\documentclass[11pt]{article}

\usepackage[margin=1in]{geometry}
\usepackage[T1]{fontenc}
\usepackage[utf8]{inputenc}
\usepackage{amsmath,amssymb}
\usepackage{booktabs}
\usepackage{graphicx}
\usepackage{microtype}
\usepackage{fancyvrb}
\usepackage[numbers,sort&compress]{natbib}
\usepackage{xcolor}
\usepackage[colorlinks=true,linkcolor=blue!60!black,citecolor=blue!60!black,urlcolor=blue!60!black]{hyperref}

\newcommand{\benchname}[1]{\textsc{#1}}
\newcommand{\condA}{A (baseline)}
\newcommand{\condB}{B (GEPA)}
\newcommand{\condC}{C (SkillOpt)}
\newcommand{\condD}{D (manual)}
\newcommand{\hashvalue}[1]{\nolinkurl{#1}}

\title{\textbf{Understanding When Reflective Prompt and Skill Optimization Helps on
Locally Served Reasoning Models: A Two-Model Study of GEPA and SkillOpt}}

\author{\parbox{0.9\textwidth}{\centering
Andrea Giummarra\textsuperscript{1}, Carmelo Baglieri\textsuperscript{1}, Antonio Di Maria\textsuperscript{1},\\
Andrea Farruggio\textsuperscript{1}, Michele Piatto\textsuperscript{1}, Giovanni Micale\textsuperscript{2}}\\[0.4em]
\small \textsuperscript{1}Evision srl, Ragusa, Italy\\
\small \textsuperscript{2}Department of Clinical and Experimental Medicine, University of Catania, Catania, Italy\\
\small Contact: \texttt{giummarra@evisionsrl.it}}

\date{October 2026}

\hypersetup{
  pdftitle={Understanding When Reflective Prompt and Skill Optimization Helps on Locally Served Reasoning Models: A Two-Model Study of GEPA and SkillOpt},
  pdfauthor={Andrea Giummarra, Carmelo Baglieri, Antonio Di Maria, Andrea Farruggio, Michele Piatto, Giovanni Micale}
}

\begin{document}

\maketitle

\begin{abstract}
Reflective text-space optimizers such as GEPA and SkillOpt have reported large
gains on frontier and mid-size models. We test replication across two
reasoning-native target-serving configurations: DeepSeek V4 Flash through
DwarfStar and Qwen3.8-27B through LM Studio. For each target, we evaluate the
same held-out tasks and four conditions - minimal baseline, GEPA prompt,
SkillOpt skill, and manually engineered prompt - on HotpotQA ($n=500$ per
condition) and MATH-500 ($n=250$). In the primary Qwen v2 rerun under the same
frozen protocol, GEPA selects the HotpotQA seed and exactly reproduces baseline
behavior, whereas SkillOpt's validation gain reverses into held-out
degradation. On MATH-500, GEPA raises
exact match from \textbf{0.828 to 0.904} while reducing latency but increasing
token consumption, so lower response time does not imply lower token cost.
DeepSeek likewise shows its strongest GEPA effect on MATH-500
\textbf{(0.852 to 0.908)}. Across both targets, GEPA's strongest gains occur on
MATH-500, while neither optimizer improves the HotpotQA baseline. Optimization
is therefore task- and method-dependent, and validation gains may reverse on
held-out evaluation. An earlier Qwen run is retained as serving-sensitivity
evidence. \textbf{These results suggest that textual headroom for a model-task
pair and the observed effect of a text-space optimizer are more informative
when decomposed into content/reasoning, answer-realization/evaluator alignment,
and serving-completion/reliability. The cross-benchmark pattern further
suggests that optimization may benefit from low-ambiguity, actionable
evaluation signals.}
\end{abstract}

\section{Introduction}
\label{sec:intro}

Large language models are increasingly adapted to downstream tasks without
touching their weights: by evolving prompts~\citep{yang2024opro,
fernando2023promptbreeder, guo2024evoprompt}, by compiling multi-stage
programs~\citep{khattab2024dspy, opsahlong2024mipro}, or by training external
textual artifacts such as skill documents~\citep{yang2026skillopt}. Two recent
systems stand out for pushing this text-space adaptation furthest. GEPA
\citep{agrawal2025gepa} evolves prompts through natural-language reflection on
execution traces and Pareto-based candidate selection, reporting gains over
reinforcement-learning post-training with a fraction of the rollouts. SkillOpt
\citep{yang2026skillopt} treats a Markdown skill file as the sole trainable
parameter of a frozen agent, accepting bounded edits only when they strictly
improve a held-out validation score, and reports best-or-tied results across
52 (model, benchmark, harness) cells.

Both papers evaluate mostly on hosted frontier or mid-size models. Meanwhile,
reasoning-native open-weight models can be served through substantially
different local stacks. We study DeepSeek V4 Flash \citep{deepseek2026v4}
through DwarfStar (ds4) \citep{sanfilippo2026ds4} and Qwen3.8-27B
\citep{qwenteam2026qwen38_27b} through LM
Studio on Apple Silicon. This setting changes the economics of text-space
optimization, but it also exposes an under-studied interaction: an artifact
may change not only the correctness of completed answers but also whether a
request completes under a particular serving stack.

We study this question through controlled within-target comparisons and a
protocol-preserving cross-target replication that preserves datasets, splits,
framework seed, conditions, metrics, and optimizer implementations.
Because the serving stacks necessarily differ, the design supports paired
within-model effects and cross-model replication analysis, not a pooled claim
that model identity alone causes every observed difference. Our research
questions are:

\begin{itemize}
  \item \textbf{RQ1.} Which GEPA and SkillOpt effects replicate across two
  reasoning-native target models under paired held-out evaluation?
  \item \textbf{RQ2.} How do content and reasoning, answer realization and
  evaluator alignment, and serving completion and reliability distinguish
  observed text-space optimization effects?
  \item \textbf{RQ3.} How strongly do optimizer effects vary across
  target-serving configurations, and how do reliability and condition order
  affect their interpretation?
  \item \textbf{RQ4.} How reliable are validation- and selection-gated
  acceptance criteria at the set sizes used here?
\end{itemize}

\paragraph{Contributions.}
(i)~A configuration-driven benchmarking framework that runs official
GEPA and SkillOpt implementations against any OpenAI-compatible endpoint,
with deterministic scoring, incremental persistence, endpoint failover, and a
statistical layer (bootstrap CIs, paired permutation tests).
(ii)~Controlled within-target comparisons and a protocol-preserving
cross-target replication spanning two models, two benchmarks, and four
conditions, with 6,000 held-out task-condition observations in total.
(iii)~A three-layer framework distinguishing content/reasoning,
answer-realization/evaluator alignment, and serving-completion/reliability,
supported by response-completion and success-conditional analyses.
(iv)~Cross-model interaction and multiplicity analyses showing HotpotQA GEPA
non-improvement across two configurations, including an exact Qwen
no-intervention outcome, task-consistent MATH GEPA gains, and method-dependent
SkillOpt transfer.
(v)~Selection-noise and optimization-trajectory analyses showing why small
adaptive validation gains should be treated as diagnostics, not deployment
guarantees.

\section{Related Work}
\label{sec:related}

\paragraph{Reflective prompt and skill optimization.}
GEPA \citep{agrawal2025gepa} samples system-level trajectories, reflects on
them in natural language, and maintains a Pareto frontier of prompt
candidates; it outperforms the GRPO reinforcement-learning baseline by up to
20\% with up to $35\times$ fewer rollouts, and also beats MIPROv2
\citep{opsahlong2024mipro}. SkillOpt \citep{yang2026skillopt} is, to our
knowledge, the first systematic text-space optimizer for agent skills: a
separate optimizer model turns scored rollouts into bounded add/delete/replace
edits on a single skill document, gated by strict improvement on a held-out
selection set, with a textual learning-rate budget and a rejected-edit buffer.
Both are used here through their official open-source implementations.

{\sloppy FLARE \citep{sundararaman2026flare} uses error-focused reflection grounded in
few-shot references; its authors report gains over GEPA across their evaluated
task--model pairs, including HotpotQA, although its retrieval-aware metric and
experimental stack preclude direct numerical comparison. WikiSkill
\citep{tang2026wikiskill} co-evolves executable skills and persistent wiki
knowledge while separating immutable experience, accumulated knowledge, and
active skills. It provides related methodological evidence rather than a
HotpotQA replication.\par}

\paragraph{Prompt-optimization lineage.}
OPRO \citep{yang2024opro} casts the LLM itself as an optimizer over prompt
space; PromptBreeder \citep{fernando2023promptbreeder} and EvoPrompt
\citep{guo2024evoprompt} apply evolutionary search; TextGrad
\citep{yuksekgonul2024textgrad} back-propagates natural-language ``gradients''
through compound systems; DSPy \citep{khattab2024dspy} compiles declarative
pipelines and optimizes their prompts jointly (MIPROv2). Several of these
methods report particularly large gains when the initial prompt is weak or
when substantial task-specific improvement remains available --- a pattern
consistent with the headroom perspective studied here.

\paragraph{Reasoning models may reduce the value of explicit prompt
engineering.}
Both targets expose a reasoning mode before the visible answer. Our HotpotQA
results (\S\ref{sec:results-hotpotqa}) show on two models that several
behaviors typically induced through extractive-QA prompting --- short-answer
discipline, span-copying, and multi-hop decomposition --- are already
competitive under a minimal instruction. This is evidence about the tested
model--task pairs, not about reasoning models as a class.

\paragraph{Validation-set overfitting.}
Our selection-noise analysis is an instance of the classic adaptive
overfitting problem: repeatedly evaluating candidates on the same small set
and keeping the best inflates the apparent score \citep{recht2019imagenet}.
The phenomenon is formalized in the adaptive data analysis literature
\citep{dwork2015stoc}, and the \emph{reusable holdout}
\citep{dwork2015reusable} was proposed precisely to preserve validity under
repeated adaptive querying of a holdout set. The statistical machinery we
use is standard: nonparametric bootstrap \citep{efron1979bootstrap} and
permutation tests in the Fisher tradition \citep{fisher1935design}.

\paragraph{Agentic design patterns.}
The Reflection and Learning-and-Adaptation patterns that GEPA and SkillOpt
instantiate are surveyed in \citet{gulli2025agentic}; we cite it as
background on the design space, not as a source on either system.

\section{Background: GEPA and SkillOpt}
\label{sec:background}

\paragraph{GEPA.}
GEPA (Genetic-Pareto) \citep{agrawal2025gepa} maintains a pool of candidate
prompts for the components of an AI system. Each iteration: (1) select a
parent candidate from the Pareto frontier of per-task validation scores; (2)
roll out the system on a minibatch of training tasks; (3) have a
\emph{reflection model} read the full trajectories (inputs, reasoning,
outputs, per-task feedback) and propose a revised prompt in natural language;
(4) evaluate the proposal on the minibatch, and, if it improves, on the full
validation set; (5) add it to the pool. The per-task (rather than aggregate)
Pareto criterion preserves candidates that win on different task subsets. The
final artifact is the candidate with the best full-validation score. The
budget is expressed in \emph{metric calls} (single-task evaluations).

\paragraph{SkillOpt.}
SkillOpt \citep{yang2026skillopt} trains a Markdown skill document that is
prepended to a frozen target model's context. Each step: (1) roll out the
target with the current skill on a training batch; (2) an \emph{analyst /
optimizer model} reflects on failures and successes and emits bounded
add/delete/replace edits (at most the \emph{textual learning rate} per step);
(3) the edited skill is evaluated on a held-out \emph{selection set}; (4) the
edit is accepted only if the selection score strictly improves over the best
so far, otherwise it enters a rejected-edit buffer that discourages
re-proposal. Deployment adds zero optimizer calls: only the final
\texttt{best\_skill.md} travels with the model.

Both methods rely, at selection time, on the assumption that improvements
measured on their validation or selection data are informative about unseen
examples. Our study examines that assumption directly.

\section{Benchmark Framework and Methodology}
\label{sec:framework}

\subsection{Architecture}

\begin{figure}[t]
  \centering
  \includegraphics[width=\linewidth]{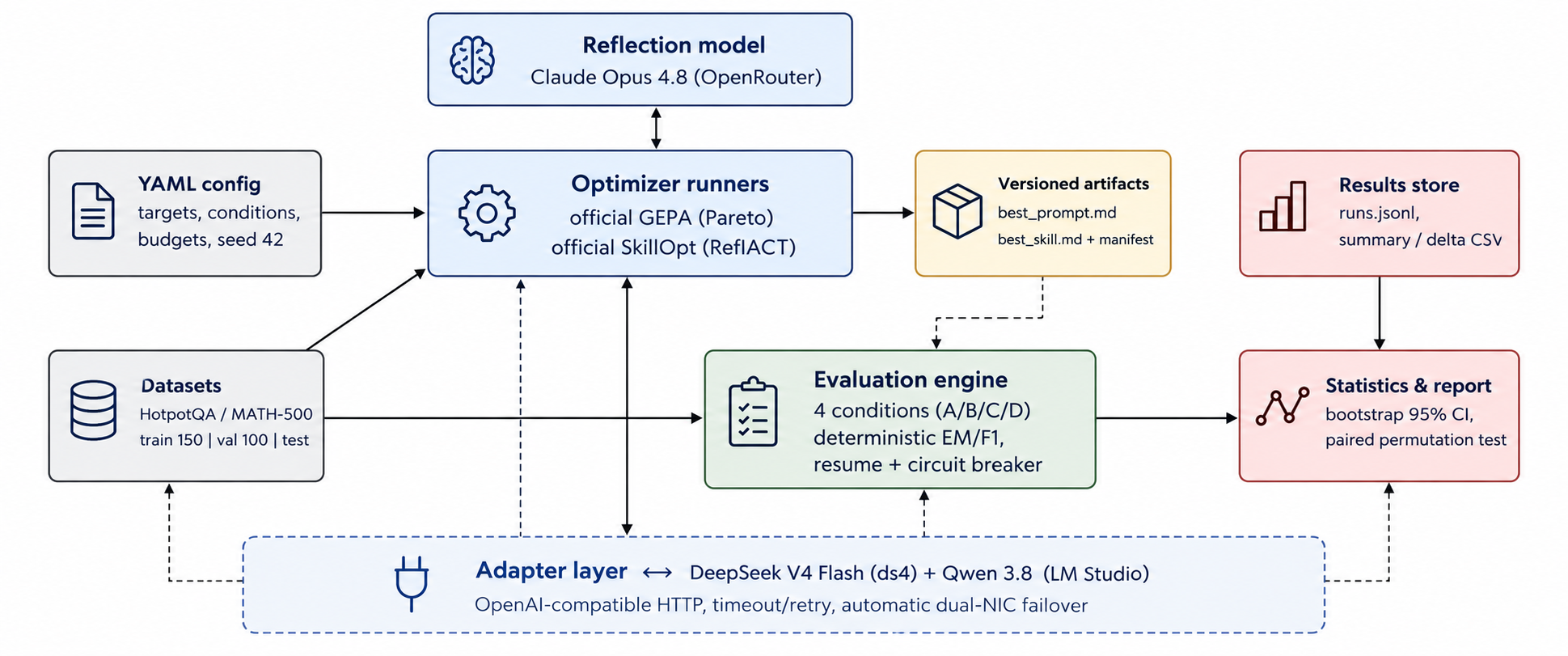}
  \caption{Framework pipeline. A single YAML configuration drives an
  optimization phase (top: optimizer runners calling the reflection model
  and emitting versioned artifacts) and an evaluation phase (bottom:
  deterministic scoring of the four conditions feeding the statistics and
  reporting layer) against one frozen target at a time; all target
  traffic passes through one OpenAI-compatible adapter with timeout, retry,
  and configured dual-NIC failover (see \S\ref{sec:framework} for the role of
  each layer). The diagram's abbreviated ``Qwen 3.8'' label denotes the
  Qwen3.8-27B target.}
  \label{fig:pipeline}
\end{figure}

The framework (Python, no heavyweight numerical dependencies) is organized in
five layers, connected as in Figure~\ref{fig:pipeline}:

\begin{enumerate}
  \item \textbf{Adapter layer.} A common interface over targets; here an
  OpenAI-compatible adapter with connection validation, per-call timeout
  control, and automatic dual-NIC endpoint failover (if the primary base URL
  stops answering, the adapter pings the alternates and retries once on the
  first reachable one).
  \item \textbf{Optimizer runners.} Thin integrations that run the
  \emph{official} \texttt{gepa} and \texttt{skillopt} packages against any
  adapter target, with archived run directories so that repeated
  optimizations never silently resume stale state.
  \item \textbf{Evaluation engine.} Deterministic per-task scoring
  (\S\ref{sec:metrics}), incremental JSONL persistence with resume support,
  and a circuit breaker that skips a target after repeated consecutive
  connection errors instead of accumulating spurious failures.
  \item \textbf{Results store and reporting.} Per-task rows
  (\texttt{runs.jsonl}), aggregated CSVs with bootstrap CIs, paired-delta
  CSVs with permutation $p$-values, and an HTML report.
  \item \textbf{Configuration layer.} A single YAML file defines targets,
  conditions, datasets, optimizer budgets, and scoring; adding a model or a
  condition is configuration, not code.
\end{enumerate}

Optimized artifacts are versioned with manifests recording seed, budgets,
reflection model, validation scores, and convergence traces; every number in
this paper is traceable to one of those files.

\subsection{The four-condition design}
\label{sec:conditions}

Both benchmarks compare four conditions on identical task sets:

\begin{itemize}
  \item \textbf{\condA:} a deliberately \emph{minimal} one-line seed prompt
  (e.g.\ ``Solve the math problem. Give the final answer inside
  \verb|\boxed{}|.'').
  \item \textbf{\condB:} the best prompt found by official GEPA, seeded from
  the minimal prompt.
  \item \textbf{\condC:} the minimal prompt plus the best skill document
  trained by official SkillOpt, seeded from a minimal one-line skill.
  \item \textbf{\condD:} a manually engineered prompt written with standard
  prompt-engineering practice, used as-is.
\end{itemize}

Accordingly, C--A estimates the deployed SkillOpt-condition effect relative
to the minimal baseline; it need not isolate accepted edits because condition C
also includes the skill wrapper even when the optimizer accepts no change.

The minimal seed is a design choice, not an oversight: in a pilot with a
hand-tuned seed (\S\ref{sec:ablation}), neither optimizer identified an
improvement over the curated seed, providing little opportunity to study
their behavior when substantial textual-context changes remain possible.
Seeding from a minimal prompt increases the
opportunity for both methods to modify the textual context; condition D
controls for the alternative of simply paying a human to write the prompt.

\subsection{Metrics and scoring}
\label{sec:metrics}

Scoring is fully deterministic (no LLM judge). For \benchname{HotpotQA} we
use SQuAD-style normalization (lowercase, strip articles and punctuation,
whitespace tokenization) and compute \emph{exact match}
$\mathrm{EM}\in\{0,1\}$ and token-level F1
\[
  \mathrm{F1} \;=\; \frac{2PR}{P+R},\qquad
  P=\frac{|\text{pred}\cap\text{gold}|}{|\text{pred}|},\qquad
  R=\frac{|\text{pred}\cap\text{gold}|}{|\text{gold}|}.
\]
For \benchname{MATH-500} we extract the last \verb|\boxed{...}| expression
(with a fallback to a ``final answer'' pattern), normalize LaTeX (fraction
shorthand, \verb|\text{}| wrappers, currency and percent symbols, spacing),
and score exact match of the normalized answers, with a numeric-equality
fallback at tolerance $10^{-6}$.

Each task also receives a format-compliance score $f$ (HotpotQA: a non-empty
answer of at most 120 characters; MATH: a parseable boxed answer) and a
robustness score $r$ ($1$ if the call succeeded, $0$ on error). The
framework's composite score is the fixed convex combination
\begin{equation}
  s \;=\; 0.45\,q + 0.20\,f + 0.20\,c + 0.15\,r,
  \label{eq:composite}
\end{equation}
where quality $q$ is F1 (HotpotQA) or EM (MATH-500) and the
exact-match/correctness component $c$ is EM. On MATH-500, $q$ and $c$ are not
independent: both contain exact-match information, so the fixed composite
intentionally gives benchmark correctness a combined weight of $0.65$.
The composite score is a framework-level operational metric rather than a
benchmark-standard metric. Accordingly, benchmark-native quality metrics ---
F1/EM for HotpotQA and EM for MATH-500 --- are treated as the primary
measures of task performance; the composite is reported as a complementary
measure that also captures format compliance and call robustness. The
weights in Eq.~\ref{eq:composite} were fixed at framework-design time,
before the reported test evaluation, and were not tuned on test performance;
their purpose is to provide a common operational score across tasks rather
than to replace benchmark-native metrics.

\subsection{Headroom}
\label{sec:headroom}

Let $s_{\text{base}}$ be the baseline score of a (model, task-distribution,
serving-stack) configuration and $s_{\max}$ its \emph{practical ceiling}:
the best end-to-end score reachable by varying only textual context while
weights, decoding parameters, and scoring remain fixed. Define the
\emph{headroom}
\begin{equation}
  H \;=\; s_{\max} - s_{\text{base}} .
  \label{eq:headroom}
\end{equation}
Any text-space optimizer's true gain is bounded by $H$. Importantly,
$s_{\max}$, and therefore $H$, is generally unobservable in a finite
experiment: observing a better prompt establishes only a lower bound on the
available headroom. Headroom should therefore be interpreted here as a
conceptual property of the (model, task, scoring) configuration rather than
as a directly estimated latent quantity. Signals consistent with
$H\approx 0$ are nonetheless observable: a strong baseline, hand-tuned
prompts that fail to improve, and optimizer validation gains that do not
transfer.

The replication motivates three analytically distinct layers. \emph{Content/
reasoning headroom} concerns substantive correctness among usable responses:
did the artifact help solve content previously answered incorrectly?
\emph{Answer-realization/evaluator-alignment headroom} concerns converting
available content into a complete, parseable, appropriately formatted final
answer. \emph{Serving-completion/reliability headroom} concerns whether an
assigned request completes and yields a response under the realized serving
conditions. These are interpretive layers rather than separately identified
causal latent variables, and the primary end-to-end estimand can mix them.
HotpotQA shows limited measurable improvement from the evaluated procedures on
either target. DeepSeek MATH primarily exhibits an
answer-realization/evaluator-alignment difference. Qwen v2 MATH instead shows
an outcome-level correctness gain among jointly successful calls at matched
baseline/GEPA completion; sequential timing and the 12 recorded timeouts remain
descriptive serving limitations rather than an explanation for that gain.

\subsection{Selection bias under repeated validation}
\label{sec:noise}

Both optimizers accept a candidate when it improves a mean score over a
validation/selection set of size $n$. For paired candidate-versus-baseline
comparisons, let $\sigma_\Delta$ denote the standard deviation across tasks
of the candidate-minus-baseline score differences; the standard error of the
mean paired gain is then $\sigma_\Delta/\sqrt{n}$. When an optimizer
evaluates $k$ candidates on the same set and keeps the maximum, then
\emph{even under the null hypothesis that no candidate is truly better than
the seed}, the expected apparent gain of the selected candidate is
approximately the expected maximum of $k$ centered Gaussians,
\begin{equation}
  \mathbb{E}\!\left[\max_{i\le k}\;\hat{\Delta}_i\right]
  \;\approx\; \frac{\sigma_\Delta}{\sqrt{n}}\,\sqrt{2\ln k},
  \label{eq:maxnoise}
\end{equation}
which grows with the number of attempts and shrinks only as $1/\sqrt{n}$.
Equation~\ref{eq:maxnoise} is intended as an order-of-magnitude heuristic
rather than a formal significance threshold. It assumes approximately
Gaussian fluctuations and abstracts away the dependence among adaptively
generated candidates; in practice, optimizer proposals are correlated and
the effective number of independent comparisons may be smaller than $k$.

We instantiate Eq.~\ref{eq:maxnoise} with $\sigma_\Delta$ computed from the
actual per-task paired deltas of each run's selected candidate against its
seed, on the same validation/selection set the gate used (GEPA per-task
validation subscores are recovered from the run state; SkillOpt selection
scores from the archived per-task evaluations). As a complementary check
that does not rely on the Gaussian approximation, we also run an
\emph{empirical null simulation}: the paired deltas are mean-centered,
resampled with random sign flips (the null of no true improvement), the best
of $k$ simulated candidates is retained, and the procedure is repeated
20{,}000 times; we report the median and 95th percentile of the resulting
best-of-$k$ apparent gain. This empirical null is likewise a reference
distribution rather than an exact model of the optimizer's adaptive
candidate-generation process; it does not preserve dependence among
successive proposals. All values are in points ($100\times$ the score):

\begin{itemize}
  \item \textbf{GEPA on HotpotQA} ($n{=}100$, $k{=}9$ full-validation
  candidates, $\sigma_\Delta{=}0.181$): approximate selection-noise scale
  $\approx 3.8$ points; empirical null best-of-9 median $2.6$, 95th
  percentile $4.6$. The observed validation gain was $+1.3$ points, well
  below this approximate selection-noise scale; considered in isolation,
  such a gain provides weak evidence of genuine transfer. It indeed failed
  to transfer (test: $-0.8$, n.s.).
  \item \textbf{SkillOpt on MATH-500} ($n{=}50$, $k{=}20$ gate evaluations,
  hard-EM gate, $\sigma_\Delta{=}0.240$): approximate selection-noise scale
  $\approx 8.3$ EM points; empirical null best-of-20 median $6.2$, 95th
  percentile $9.8$. The observed selection staircase of $+6.0$ EM sits
  \emph{below the median} of the null distribution; on test it collapsed to
  $+0.4$ paired EM (n.s.).
  \item \textbf{SkillOpt on HotpotQA} ($n{=}100$, $k{=}20$ gate evaluations,
  soft-F1 gate, $\sigma_\Delta{=}0.323$): approximate selection-noise scale
  $\approx 7.9$ F1 points; empirical null median $5.9$, 95th percentile
  $9.1$. The single accepted edit improved selection F1 by only $+0.8$
  points --- and degraded test F1 by $6.2$ points.
  \item \textbf{GEPA on MATH-500} ($n{=}100$, $k{=}3$ full-validation
  candidates, $\sigma_\Delta{=}0.231$): approximate selection-noise scale
  $\approx 3.4$ points; empirical null best-of-3 median $1.9$, 95th
  percentile $4.9$. The observed $+6.0$ validation gain exceeds this
  heuristic scale and the 95th percentile of the empirical null, and, in
  this run, also transferred to the held-out test set ($+4.36$ composite,
  $p{=}0.0019$).
\end{itemize}

Qwen v2 further weakens a simple threshold interpretation. HotpotQA GEPA
selected the seed itself. HotpotQA SkillOpt accepted one of 20 edits and
raised selection F1 from $0.620$ to $0.776$, yet held-out F1 fell by 3.22
points. On MATH-500, GEPA raised validation composite from $0.7745$ to
$0.8250$ and held-out EM by 7.6 points at matched baseline/GEPA completion;
SkillOpt raised selection EM from $0.72$ to $0.78$ and held-out EM by 3.6
points while recording nine timeouts. Because no v2 empirical-null
distribution was persisted, we do not compare these gains with the inherited
DeepSeek null percentiles. We use Eq.~\ref{eq:maxnoise} only as a diagnostic
of how weak small adaptive validation gains may be; it is not a validated
acceptance threshold and it does not model serving reliability.

\subsection{Statistical protocol}
\label{sec:stats}

Per-condition means are reported with nonparametric bootstrap 95\% confidence
intervals \citep{efron1979bootstrap} (resampling tasks, $\geq$5{,}000
replicates). Condition comparisons use a \emph{paired} sign-permutation test
\citep{fisher1935design}: for conditions $X,Y$ evaluated on the same tasks,
the null hypothesis is that the per-task deltas $d_t = s_Y(t)-s_X(t)$ are
symmetric around zero; we randomly flip the signs of the nonzero deltas
($10^4$ permutations) and report the two-sided tail probability of the
observed sum. Pairing on identical task sets removes between-task variance,
which dominates in QA/math benchmarks, and thus gives far more power than
unpaired comparisons at the same $n$.

All primary hypothesis tests compare each optimized condition with the
minimal baseline on the benchmark-native quality metric: B~vs.~A and
C~vs.~A on F1 for HotpotQA, and B~vs.~A and C~vs.~A on EM for MATH-500.
Other pairwise comparisons are treated as exploratory and reported with
unadjusted $p$-values unless otherwise stated. As a familywise sensitivity
analysis, we apply Holm's procedure to the eight primary B--A and C--A tests
across two models and two benchmarks. We also estimate exploratory
model-by-condition interactions as differences between the within-model
paired effects, using task-level bootstrap intervals and sign-permutation
tests. These interactions mix target-model and serving-stack differences and
are interpreted as heterogeneity, not as the causal effect of model identity.
For Qwen MATH, task-level intervals and permutation $p$-values are conditional
on the realized serving trajectory and do not capture uncertainty induced by
temporally correlated endpoint failures.

\section{Experimental Setup}
\label{sec:setup}

\subsection{Datasets and splits}

\benchname{HotpotQA} \citep{yang2018hotpotqa}: multi-hop QA in the distractor
setting (10 paragraphs, 2 gold + 8 distractors). Examples are drawn from the
official development set of the distractor setting --- the HuggingFace
\texttt{hotpot\_qa} dataset, config \texttt{distractor}, split
\texttt{validation}, containing 7,405 examples (file \texttt{hotpot\_dev\_\allowbreak
distractor\_v1}) --- before deterministic shuffling and partitioning.
\benchname{MATH-500} \citep{lightman2024verify}: the 500-problem
competition-math subset of MATH \citep{hendrycks2021math} popularized by
\citet{lightman2024verify}, from \texttt{HuggingFaceH4/MATH-500}. Both are
shuffled with a fixed seed (42) and split sequentially: train 150 /
validation 100 / test 500 (HotpotQA) and train 150 / validation 100 / test
250 (MATH-500). Thus the ``test'' subsets reported in this paper refer to
our held-out partitions of these datasets, not to additional official
benchmark splits. Sequential slicing guarantees that enlarging the test set
never touches train/val. GEPA uses the train split for reflection
minibatches and the validation split for candidate scoring; SkillOpt uses
train for rollouts and a selection subset of the validation split (HotpotQA
100, MATH 50) for gating. The test split was not used for optimizer
candidate selection or artifact construction.

\subsection{Target models and serving stacks}
\label{sec:hardware}

The first target is \textbf{DeepSeek V4 Flash} \citep{deepseek2026v4}, a
284B-parameter mixture-of-experts model with 13B activated parameters and an
explicit thinking mode, served locally by \textbf{DwarfStar (ds4)}
\citep{sanfilippo2026ds4}: a native C inference engine with a Metal backend
and an extremely asymmetric 2/8-bit quantization recipe (checkpoint
\texttt{DeepSeek-V4-Flash-IQ2XXS-w2Q2K-\allowbreak AProjQ8-SExpQ8-OutQ8};
routed experts
at $\sim$2 bits, attention/shared/output paths at 8 bits), 250k-token
context, and a disk-backed KV cache behind an OpenAI-compatible HTTP server.
{\sloppy For reproducibility, the exact model artifact was
\texttt{DeepSeek-V4-Flash-IQ2XXS-w2Q2K-AProjQ8-\allowbreak%
SExpQ8-OutQ8-chat-v2-imatrix-0731.gguf} (SHA-256
\hashvalue{ca22ae2f838e14077c22bc1c1417b71b45b5e5a3687bd96c2ac6e17fdb6261c0}), obtained as the \texttt{ds4f-q2} download
target of DwarfStar's \texttt{download\_model.sh} from the
\texttt{antirez/deepseek-v4-gguf} Hugging Face repository
\citep{sanfilippo2026ds4}; the same server build, checkpoint, and
thinking-mode configuration were used throughout all conditions.\par} Hardware: 14-inch MacBook
Pro (November 2023), Apple M3 Max, 128\,GB unified memory, macOS Sequoia
15.7.7. The benchmark client runs on a separate Mac over LAN with automatic
dual-NIC failover. Decoding: temperature $0$, \texttt{max\_tokens} 512
(HotpotQA) / 4096 (MATH), timeout 180\,s (HotpotQA) / 480\,s (MATH), one
retry per task. Generated-token counts reported in \S\ref{sec:results} are
the \texttt{completion\_tokens} usage field returned by the server and
include the model's internal reasoning trace as well as the visible final
response.

{\sloppy The replication target is \textbf{Qwen3.8-27B}
\citep{qwenteam2026qwen38_27b}, frozen as the 8-bit MLX
variant \texttt{qwen/qwen3.8-27b@8bit} (29,528,355,154 bytes), served by LM
Studio 0.4.14+4 on a separate Apple M3 Max system with 128\,GB unified memory
and macOS 15.7.7. The loaded context was 32,768 tokens and parallelism was
one. The OpenAI-compatible endpoint reported model ID
\texttt{qwen/qwen3.8-27b}; a live preflight required both hidden reasoning
content and reasoning-token usage fields. Reasoning used the frozen LM Studio
server default (\texttt{on}); request-level reasoning-effort and
thinking-preservation controls were neither sent nor recorded. The exact local snapshot
is identified by per-file SHA-256 hashes in the retained manifest; its upstream
revision, MLX conversion revision/tool, and monolithic archive hash were not
recorded. Temperature and maximum completion tokens matched the DeepSeek
experiment; request timeouts were 180\,s for
HotpotQA and 480\,s for MATH-500, with one retry. The primary and failover
endpoint configuration, checkpoint metadata, hashes, and preflight are recorded
in the frozen replication manifests. Absolute latency and token counts are
not compared across targets because engines, quantizations, and serving hosts
differ.\par}

The Qwen experiment was a frozen target substitution: task files and ordered
splits were byte-identical to the reference experiment, and the seed,
conditions, scoring, optimizer packages, reflection model, and optimizer/test
boundary were preserved. Qwen v2 repeated the complete optimization and
evaluation pipeline in a separate namespace; it selected new artifacts in
three of the four optimizer--benchmark cells. The same test items had already
been evaluated in v1, but remained mechanically excluded from v2 optimization,
and v2 artifacts were frozen before evaluation. Conditions were evaluated
sequentially, so time-varying server availability can still be associated with
condition. We therefore prioritize paired within-target effects and treat
cross-target comparisons as descriptive replication evidence rather than a
controlled hardware- or model-only contrast.

\subsection{Optimizer configuration}

Both optimizers use \textbf{Claude Opus 4.8} (via OpenRouter) as the
reflection/optimizer model. Each optimization run keeps its target fixed to
the corresponding DeepSeek or Qwen model.
GEPA: \texttt{max\_metric\_calls} 1200 (HotpotQA) / 400 (MATH-500),
reflection minibatch 5, official Pareto candidate selection. SkillOpt
(official ReflACT trainer): 2 epochs $\times$ 10 steps, batch size 15,
textual learning rate (edit budget) 4, selection set 100 with soft-F1 gate
(HotpotQA) / 50 with hard-EM gate (MATH-500), reflect minibatch 8. Dataset
splitting, optimization scheduling, and the reported resampling procedures used
framework seed 42. Target requests did not include a provider seed, so
provider-side sampling determinism is not claimed. Optimizer budgets and gate
metrics differ across benchmarks
because they were configured according to task-specific evaluation
characteristics; consequently, cross-benchmark comparisons are descriptive
rather than controlled comparisons of benchmark identity alone. The official
GEPA engine checks its stopping condition only at iteration boundaries, and
an accepted minibatch improvement triggers a full validation evaluation that
completes regardless of remaining budget; consequently, the nominal budgets
of 1200 and 400 resulted in 1220 and 445 recorded metric calls,
respectively. The primary Qwen v2 runs used the same nominal budgets and
stopping semantics. Their realized target metric calls were 1220 and 430 for
GEPA on HotpotQA and MATH-500; SkillOpt made 67 and 102 recorded target
rollouts and accepted 1 of 20 candidate edits on HotpotQA and 2 of 20 on
MATH-500. These telemetry counts follow
the frozen manifests; they are not interchangeable with the larger number of
per-example target invocations implied by a SkillOpt batch.

\paragraph{Failure scoring and Qwen execution policy.}{\sloppy
A failed call receives zero on the benchmark-native and composite outcomes,
so the primary estimand is end-to-end performance under the observed serving
process. The primary Qwen v2 evaluation ran once without resume,
circuit-breaker activation, recovery, or post-hoc tuning. HotpotQA contains
2,000 successful unique rows. MATH-500 contains 1,000 unique rows and retains
12 read timeouts (A/B/C/D: 1/1/9/1) as zero-scored outcomes. The earlier v1
execution, including its authorized recovery after a circuit-breaker event,
is preserved separately and analyzed in \S\ref{sec:qwen-sensitivity}; it is
not pooled with v2.\par}

\section{Results}
\label{sec:results}

\begin{figure}[t]
  \centering
  \includegraphics[width=\linewidth]{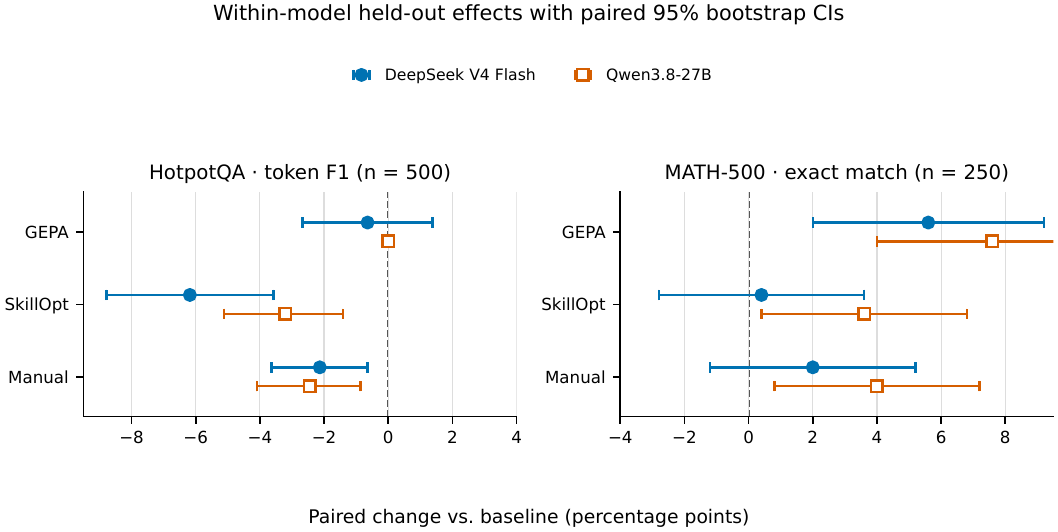}
  \caption{Paired changes from the minimal baseline in the benchmark-native
  primary metric, with task-bootstrap 95\% confidence intervals. Each panel
  uses the same held-out tasks within a model (HotpotQA $n{=}500$; MATH-500
  $n{=}250$). The forest plot visualizes within-model effects; it does not
  treat the different serving stacks as exchangeable.}
  \label{fig:effects}
\end{figure}

Table~\ref{tab:overview} and Fig.~\ref{fig:effects} summarize the primary
end-to-end results. Failed calls score zero, as specified before evaluation.
Consequently, these outcomes measure the deployed model--prompt--server
system rather than correctness conditional on receiving an answer.

\begin{table}[t]
  \centering
  \caption{Held-out benchmark-native score by target and condition. Values
  in brackets are bootstrap 95\% CIs; $\Delta$ and raw paired permutation
  $p$ compare each condition with A. OK is the number of successful calls.
  Scores and intervals use four decimal places; deltas are percentage points.
  Bold marks the best primary score within each target--benchmark block (ties
  included).
  Holm-adjusted results are reported in \S\ref{sec:crossmodel}.}
  \label{tab:overview}
  \resizebox{\textwidth}{!}{%
  \begin{tabular}{lllrlll}
    \toprule
    Benchmark & Target & Condition & Score [95\% CI] & $\Delta$ vs A (pp) & Raw $p$ & OK/$n$ \\
    \midrule
    HotpotQA & DeepSeek & A baseline & \textbf{0.8150 [0.7857, 0.8436]} & --- & --- & 500/500 \\
              &          & B GEPA & 0.8086 [0.7795, 0.8381] & $-0.64$ & .5363 & 499/500 \\
              &          & C SkillOpt & 0.7532 [0.7196, 0.7867] & $-6.18$ & $<.0001$ & 500/500 \\
              &          & D manual & 0.7937 [0.7634, 0.8243] & $-2.13$ & .0036 & 500/500 \\
              & Qwen & A baseline & \textbf{0.8215 [0.7918, 0.8491]} & --- & --- & 500/500 \\
              &      & B GEPA & \textbf{0.8215 [0.7918, 0.8491]} & $0.00$ & 1.0000 & 500/500 \\
              &      & C SkillOpt & 0.7893 [0.7561, 0.8204] & $-3.22$ & .0004 & 500/500 \\
              &      & D manual & 0.7971 [0.7652, 0.8270] & $-2.44$ & .0024 & 500/500 \\
    \addlinespace
    MATH-500 & DeepSeek & A baseline & 0.8520 [0.8080, 0.8920] & --- & --- & 250/250 \\
             &          & B GEPA & \textbf{0.9080 [0.8720, 0.9400]} & $+5.60$ & .0047 & 250/250 \\
             &          & C SkillOpt & 0.8560 [0.8120, 0.9000] & $+0.40$ & 1.0000 & 249/250 \\
             &          & D manual & 0.8720 [0.8280, 0.9120] & $+2.00$ & .3055 & 250/250 \\
             & Qwen & A baseline & 0.8280 [0.7800, 0.8760] & --- & --- & 249/250 \\
             &      & B GEPA & \textbf{0.9040 [0.8640, 0.9400]} & $+7.60$ & .0002 & 249/250 \\
             &      & C SkillOpt & 0.8640 [0.8200, 0.9040] & $+3.60$ & .0463 & 241/250 \\
             &      & D manual & 0.8680 [0.8240, 0.9080] & $+4.00$ & .0302 & 249/250 \\
    \bottomrule
  \end{tabular}
  }
\end{table}

\subsection{DeepSeek replication reference}

We first retain the detailed DeepSeek analysis because it establishes the
reference effects that Qwen is intended to replicate. The tables in the next
two subsections are DeepSeek-only; cross-model conclusions follow in
\S\ref{sec:crossmodel}.

\subsection{DeepSeek HotpotQA: no measurable improvement and a harmful accepted edit}
\label{sec:results-hotpotqa}

Table~\ref{tab:hotpotqa} reports the held-out test results ($n{=}500$).

\begin{table}[t]
  \centering
  \caption{DeepSeek \benchname{HotpotQA} test results ($n{=}500$ tasks). F1 is the
  benchmark-native primary metric; composite is Eq.~\ref{eq:composite}.
  Deltas are paired vs.\ condition A over all $n$ tasks with
  sign-permutation $p$-values (in points); a failed call scores 0. Negative
  deltas are regressions. Bold marks the column-best score, format rate, or
  latency (lower is better); p-values communicate significance separately. No
  optimizer delta versus A constitutes a significant improvement.}
  \label{tab:hotpotqa}
  \small
  \setlength{\tabcolsep}{5pt}
  \begin{tabular}{lcccccc}
    \toprule
    Condition & F1 & EM & Composite [95\% CI] & Format & Lat.\ (s) &
    $\Delta$F1 vs A ($p$) \\
    \midrule
    A baseline (minimal) & \textbf{0.8150} & 0.6720 & \textbf{0.8456} [0.8233, 0.8664] & \textbf{0.9720} & \textbf{15.2} & --- \\
    B GEPA               & 0.8086 & \textbf{0.6800} & 0.8376 [0.8133, 0.8605] & 0.9400 & 21.7 & $-0.64$ (0.5363) \\
    C SkillOpt           & 0.7532 & 0.6300 & 0.7885 [0.7609, 0.8154] & 0.8680 & 23.0 & $-6.18$ ($<10^{-4}$) \\
    D manual             & 0.7937 & 0.6600 & 0.8260 [0.8016, 0.8494] & 0.9340 & 18.9 & $-2.13$ (0.0036) \\
    \bottomrule
  \end{tabular}
\end{table}

Three findings. First, \emph{the minimal one-line prompt is the strongest
tested condition on the held-out set}: the hand-engineered prompt (D) is
significantly worse than the minimal seed (F1 $-2.13$, $p{=}0.0036$;
composite $-1.96$, $p{=}0.0015$), and GEPA's evolved prompt is statistically
indistinguishable from the seed (F1 $-0.64$, $p{=}0.54$; EM $+0.8$,
$p{=}0.70$). In this setting, the reasoning-native model already exhibits
much of the span-extraction and short-answer behavior targeted by
conventional QA instructions; the additional instructions do not improve
held-out quality and increase latency (15.2\,s $\to$ 18.9--23.0\,s, with
mean output length growing from 111 to 153--218 tokens).

Second, \emph{SkillOpt's single accepted edit actively hurts}. Under the
soft-F1 gate, 1 of 20 candidate edits was accepted, at training step 18 ($+0.8$ F1 on
the selection set); on test it costs $-6.2$ F1 and $-4.2$ EM
($p{\le}0.02$), and drops format compliance from $0.972$ to $0.868$. The
edit instructs the model to include attached qualifiers and phrases in its
answers (\S\ref{app:artifacts}); this systematically lengthens answers,
which exact-match and F1 scoring punish. A verbosity-inducing rule slipped
through a gate that a $+0.8$-point selection improvement --- far inside the
$\approx$7.9-point approximate selection-noise scale of
Eq.~\ref{eq:maxnoise} --- could not justify.

\begin{figure}[t]
  \centering
  \includegraphics[width=\linewidth]{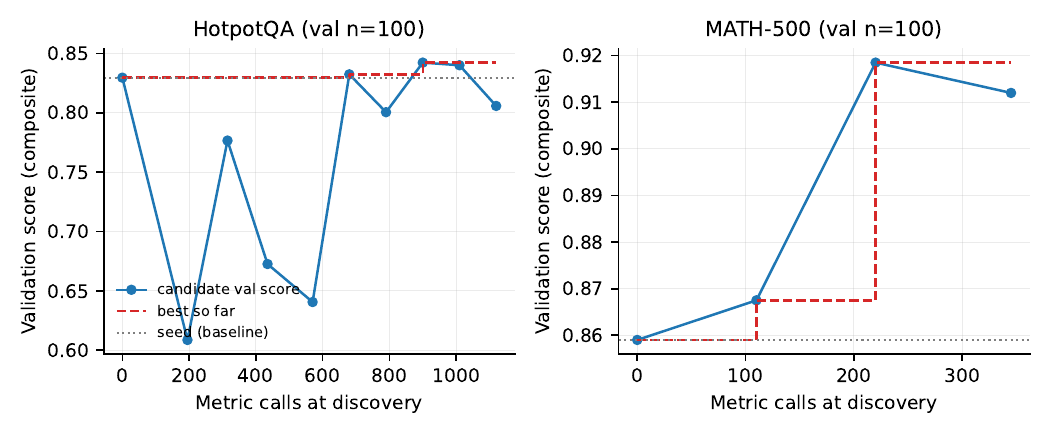}
  \caption{DeepSeek GEPA candidate validation scores vs.\ metric-call budget. Left
  (HotpotQA): candidates scatter mostly below the seed; the best-so-far line
  creeps up by $+1.3$ points, below the approximate selection-noise scale.
  Right (MATH-500): the deployed best candidate is discovered by 220 metric
  calls, with a $+6.0$-point climb beyond that scale; the complete optimization
  run consumes 445 metric calls, and the gain transfers to test.}
  \label{fig:gepa}
\end{figure}

Third, \emph{GEPA's validation gain did not transfer}. The validation
composite increased from $0.830$ to $0.842$ ($+1.3$ points), a gain small
relative to the approximate selection-noise scale of $\approx$3.8 points
discussed in \S\ref{sec:noise} (Fig.~\ref{fig:gepa} left shows candidates
mostly \emph{far below} the seed); on the held-out test set the selected
prompt was $-0.8$ composite points below baseline.

\subsection{DeepSeek MATH-500: significant GEPA gains}
\label{sec:results-math}

\begin{table}[t]
  \centering
  \caption{DeepSeek \benchname{MATH-500} test results ($n{=}250$ tasks). EM is the
  benchmark-native primary metric; composite is Eq.~\ref{eq:composite}.
  Deltas are paired vs.\ condition A over all $n$ tasks with
  sign-permutation $p$-values (in points); a failed call scores 0. GEPA is
  the only condition with a significant improvement; it also dominates on
  latency. Bold marks the column-best score, format rate, or latency (lower is
  better); p-values communicate significance separately.}
  \label{tab:math}
  \small
  \begin{tabular}{lccccc}
    \toprule
    Condition & EM & Composite [95\% CI] & Format & Lat.\ (s) &
    $\Delta$EM vs A ($p$) \\
    \midrule
    A baseline (minimal) & 0.8520 & 0.8846 [0.8480, 0.9188] & 0.9040 & 65.9 & --- \\
    B GEPA               & \textbf{0.9080} & \textbf{0.9282} [0.8982, 0.9554] & \textbf{0.9400} & \textbf{45.8} & $+5.60$ (0.0047) \\
    C SkillOpt           & 0.8560 & 0.8866 [0.8508, 0.9198] & 0.9040 & 64.9 & $+0.40$ (1.0000) \\
    D manual             & 0.8720 & 0.9016 [0.8676, 0.9332] & 0.9240 & 57.4 & $+2.00$ (0.3055) \\
    \bottomrule
  \end{tabular}
\end{table}

\begin{figure}[t]
  \centering
  \includegraphics[width=\linewidth]{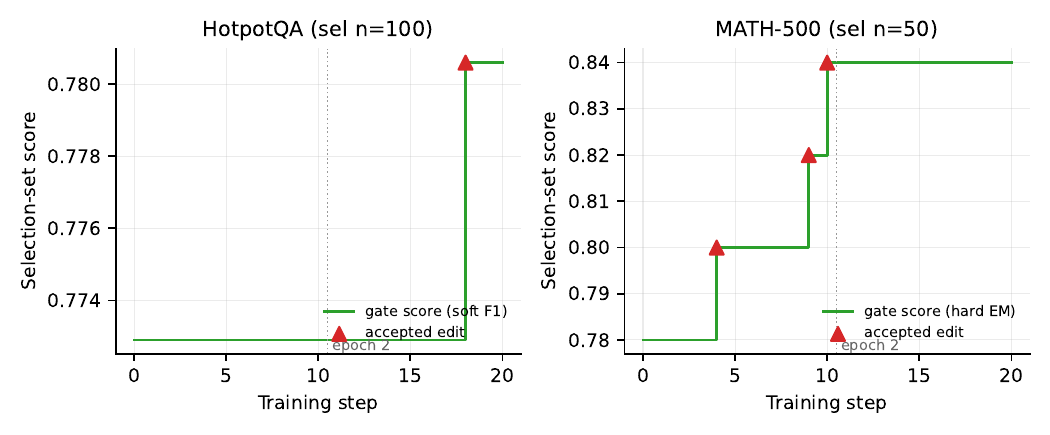}
  \caption{DeepSeek SkillOpt gate trajectory (selection-set score per training step;
  triangles mark accepted edits). Right (MATH-500, hard-EM gate, $n{=}50$): a
  $0.78\to0.84$ staircase in epoch 1, plateau in epoch 2. Left (HotpotQA,
  soft-F1 gate, $n{=}100$): flat except one late accept --- the edit that
  degrades test F1 by 6.2 points.}
  \label{fig:skillopt}
\end{figure}

Table~\ref{tab:math} shows the mirror image. GEPA improves exact match from
$0.852$ to $0.908$ ($+5.6$ EM, paired $p{=}0.0047$; composite $+4.36$,
$p{=}0.0019$). Exploratory pairwise comparisons also favored GEPA over
SkillOpt ($+4.16$ composite, unadjusted $p{=}0.011$) and over the manual
condition ($+2.66$, unadjusted $p{=}0.047$). SkillOpt's selection-set
staircase ($0.78\to0.84$ hard EM, 3 of 20 candidate edits accepted;
Fig.~\ref{fig:skillopt} right) collapses on test to a paired $+0.4$ EM
($p{=}1.0$). Manual prompt engineering lands in between and is not
significant ($+2.0$ EM, $p{=}0.3055$). These results establish held-out
generalization of the artifact obtained in this optimization run; they do
not by themselves establish the run-to-run stability of GEPA.

\begin{figure}[t]
  \centering
  \includegraphics[width=\linewidth]{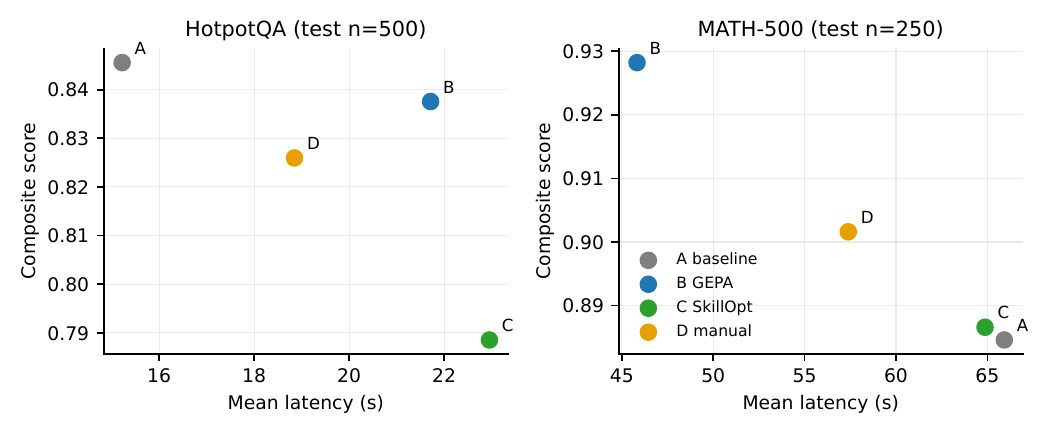}
  \caption{DeepSeek quality vs.\ mean latency. On MATH-500 (right), GEPA
  Pareto-dominates every other condition: higher composite score at the
  lowest latency. On HotpotQA (left), the baseline dominates: every longer
  prompt costs latency without quality.}
  \label{fig:pareto}
\end{figure}

\paragraph{Pareto dominance on quality and latency.}
Observed mean latency is 30\% lower under GEPA ($65.9\to45.8$\,s) \emph{while}
increasing benchmark EM (Fig.~\ref{fig:pareto}). The mechanism is visible in the
artifact (\S\ref{app:artifacts}) and in the output statistics: the evolved
prompt orders the model to emit \emph{only} the boxed answer, suppressing
post-reasoning prose --- mean generated output drops from 1104 tokens
(median 515) under the baseline to 727 (median 263) under GEPA --- and pins
down gold-format conventions (exact values instead of decimals;
multiple-choice answers as \verb|\boxed{\text{(C)}}|) that turn near-misses
into exact matches. Relative to the $14.8$-point gap between the baseline EM
and the metric upper bound of $1.0$, the observed GEPA difference accounts for
$5.6$ points, or approximately $38\%$ of that observable gap, in
445~metric calls.

\begin{figure}[t]
  \centering
  \includegraphics[width=0.78\linewidth]{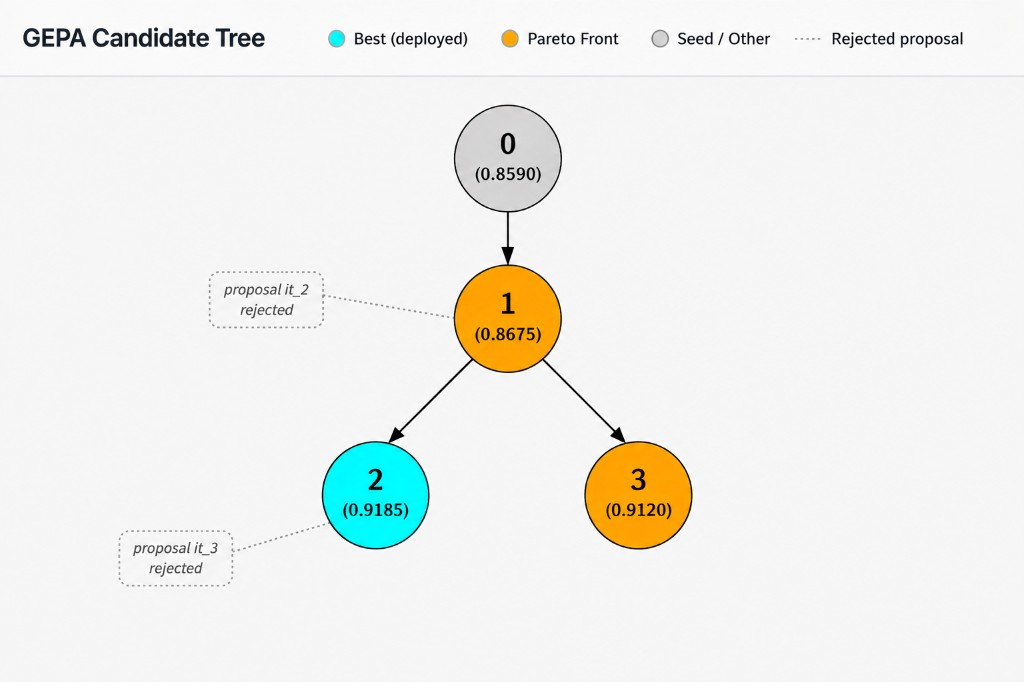}
  \caption{GEPA candidate tree on MATH-500, redrawn from the run's own
  \texttt{candidate\_tree.html} and \texttt{run\_log.json}. Solid circles are
  candidates accepted into the pool, labeled with their full-validation
  scores ($n{=}100$); arrows point from parent to child; dotted callouts are
  proposals rejected at the reflection-minibatch stage. Candidate 2 (cyan) is
  the deployed best; candidate 3 remains on the per-task Pareto front
  (orange) despite a lower aggregate score.}
  \label{fig:tree}
\end{figure}

\paragraph{How GEPA got there: the candidate tree.}
Figure~\ref{fig:tree} reconstructs the full evolutionary history of the
winning MATH-500 prompt from GEPA's own run telemetry. In this run, the
search trajectory is short
and directed: from the one-line seed (candidate 0, val $0.859$), the first
reflection produced candidate 1 ($0.8675$), which introduced the boxed-only
output rule and the exact-values-over-decimals convention; a second
reflection on candidate 1 produced candidate 2 ($0.9185$), which added the
multiple-choice \verb|\boxed{\text{(C)}}| convention and the domain reminders
and became the deployed artifact. Two further
proposals (iterations 2 and 3) failed their reflection minibatch and were
discarded without a full validation pass --- GEPA's cheap early rejection at
work --- and a last accepted branch (candidate 3, $0.912$) improved on some
validation tasks but not in aggregate, so it stayed on the Pareto front
without displacing candidate 2. The tree makes GEPA's two selection
mechanisms visible at once: minibatch gating prunes bad proposals early,
while the per-task Pareto front keeps candidate 3 alive as a possible parent
even though its aggregate score is lower.

\subsection{Qwen v2: exact HotpotQA no-intervention, SkillOpt reversal, and MATH gains}
\label{sec:qwen}

On HotpotQA, Qwen v2 GEPA selected the one-line seed. The baseline and GEPA
artifacts have the same hash, and all 500 paired responses are identical (F1
$0.8215$ under both conditions). This is an exact no-intervention outcome of
the realized search, not evidence that GEPA cannot improve HotpotQA in
general. SkillOpt exhibits the opposite regime: its validation score rises
from $0.620$ to $0.776$, but held-out F1 falls from $0.8215$ to $0.7893$
($\Delta=-0.0322$, paired 95\% CI $[-0.0512,-0.0142]$, $p{=}.0004$).
Baseline alone solves 14 paired items and SkillOpt alone solves 10; format
compliance falls from $0.940$ to $0.884$, mean latency rises by 25.7\%, and
observed total tokens rise from 832,066 to 957,880. The manual prompt also
degrades F1 ($\Delta=-0.0244$, $p{=}.0024$). Thus the large optimization-time
SkillOpt gain reverses on the held-out panel.

\begin{figure}[t]
  \centering
  \includegraphics[width=\linewidth]{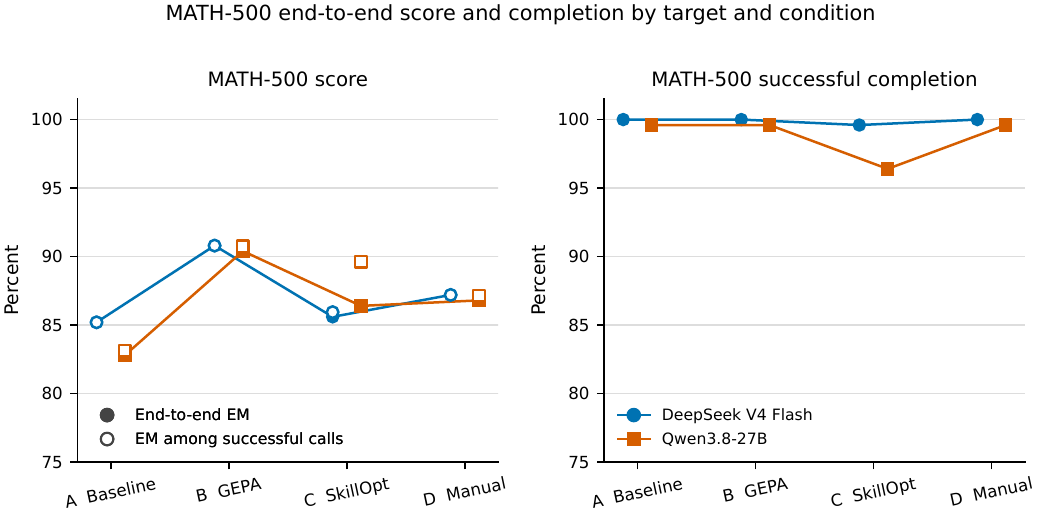}
  \caption{MATH-500 primary end-to-end EM (left) and successful-call rate
  (right) for both DeepSeek and Qwen, each over all 250 assigned tasks. Qwen v2 completion is 99.6\%
  under A, B, and D and 96.4\% under C; the GEPA score gain therefore remains
  visible at matched completion. Cross-target absolute latency is omitted
  because serving stacks differ.}
  \label{fig:mathreliability}
\end{figure}

On MATH-500, GEPA improves EM from $0.828$ to $0.904$
($\Delta=+0.076$, paired 95\% CI $[0.040,0.116]$, $p{=}.0002$). Of the 250
paired items, 204 are correct under both conditions, 22 only under GEPA, 3
only under baseline, and 21 under neither. Both conditions record one timeout,
so the observed gain is not produced by differential completion coverage.
Figure~\ref{fig:mathreliability} and Table~\ref{tab:reliability} separate the
quality difference from completion coverage.
Mean end-to-end latency is lower under GEPA (143.34\,s to 99.35\,s,
$-30.7\%$), although conditions ran sequentially and latency was not subjected
to an inferential test. SkillOpt reaches EM $0.864$ ($\Delta=+0.036$, 95\% CI
$[0.004,0.068]$, $p{=}.0463$) and records nine timeouts; its raw paired effect
is smaller and borderline and does not survive Holm correction. The manual
condition reaches $0.868$
($\Delta=+0.040$, exploratory $p{=}.0302$).

\begin{table}[t]
  \centering
  \caption{Qwen v2 MATH reliability decomposition ($n{=}250$ per condition).
  End-to-end EM scores failures as zero. EM$\mid$OK conditions on successful
  calls; paired OK $\Delta$ restricts each comparison to tasks where both A
  and the comparator succeeded and is descriptive. Bold marks the column-best
  value (ties included).}
  \label{tab:reliability}
  \small
  \begin{tabular}{lrrrr}
    \toprule
    Condition & End-to-end EM & OK/$n$ & EM$\mid$OK & Paired OK $\Delta$ vs A \\
    \midrule
    A baseline & 0.8280 & \textbf{249/250} & 0.8313 & --- \\
    B GEPA     & \textbf{0.9040} & \textbf{249/250} & \textbf{0.9076} & \textbf{$+0.0726$} \\
    C SkillOpt & 0.8640 & 241/250 & 0.8963 & $+0.0375$ \\
    D manual   & 0.8680 & \textbf{249/250} & 0.8715 & $+0.0363$ \\
    \bottomrule
  \end{tabular}
\end{table}

\begin{table}[t]
  \centering
  \caption{Recorded Qwen v2 MATH failure categories in the final assigned-task
  panel. The uninterrupted evaluation records only read timeouts and required
  no recovery. Bold marks the column minimum (ties included).}
  \label{tab:qwenfailures}
  \small
  \begin{tabular}{lrrr}
    \toprule
    Condition & Read timeout & Other errors & Final failures \\
    \midrule
    A baseline & \textbf{1} & \textbf{0} & \textbf{1} \\
    B GEPA     & \textbf{1} & \textbf{0} & \textbf{1} \\
    C SkillOpt & 9 & \textbf{0} & 9 \\
    D manual   & \textbf{1} & \textbf{0} & \textbf{1} \\
    \bottomrule
  \end{tabular}
\end{table}

\begin{figure}[t]
  \centering
  \includegraphics[width=\linewidth]{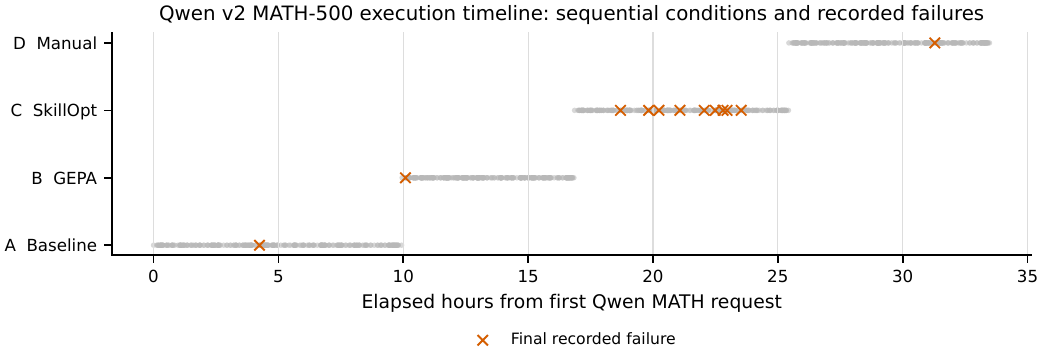}
  \caption{Qwen v2 MATH-500 execution timeline from retained per-task
  timestamps. Conditions ran sequentially; crosses mark the 12 recorded read
  timeouts. The figure diagnoses temporal non-exchangeability and does not
  attribute failures causally to prompt text.}
  \label{fig:qwentimeline}
\end{figure}

The failure categories in Table~\ref{tab:qwenfailures} and their temporal
placement in Figure~\ref{fig:qwentimeline} make the sequential serving
trajectory explicit.

Difficulty-stratified EM is exploratory. On 139 difficult items, A, B, C,
and D score 0.741, 0.871, 0.791, and 0.806; on 46 medium items they score
0.935, 0.957, 0.978, and 0.935; and on 65 easy items they score
0.938, 0.938, 0.938, and 0.954. The largest descriptive GEPA difference occurs in
intermediate algebra ($n{=}52$, 0.635 to 0.923). These subgroup estimates
were not separately powered or tested and should not be treated as
confirmatory.

\begin{figure}[t]
  \centering
  \includegraphics[width=\linewidth]{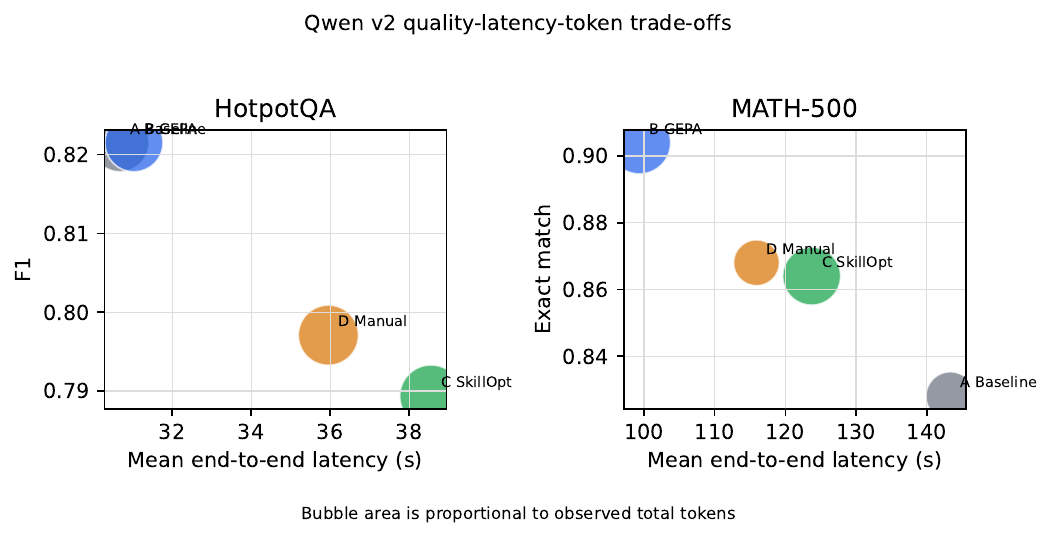}
  \caption{Qwen v2 quality, mean end-to-end latency, and observed total-token
  trade-offs. Bubble area is proportional to total reported tokens. Within a
  benchmark, lower latency and higher score are preferable, but smaller
  bubbles represent a distinct token-economy objective.}
  \label{fig:qwen-tradeoffs}
\end{figure}

Figure~\ref{fig:qwen-tradeoffs} makes the resource trade-off explicit. On
MATH-500, GEPA produces fewer output and reasoning tokens than baseline, but
its longer prompt increases input tokens from roughly 163 to 1,414 and total
tokens per successful response from 1,382 to 2,253 ($+63\%$). Hence it is
faster in observed wall time but less token-efficient. Manual prompting is
the most economical MATH condition by total tokens, while GEPA yields the
highest accuracy and lowest observed latency.

\subsection{Cross-model replication, heterogeneity, and multiplicity}
\label{sec:crossmodel}

The strongest cross-model pattern is task-level. GEPA does not improve
HotpotQA on either target: DeepSeek changes by $-0.64$ F1 points and Qwen v2
selects the seed, producing an exact zero effect. SkillOpt is harmful in both
realized HotpotQA runs ($-6.18$ DeepSeek; $-3.22$ Qwen), although the
Qwen-minus-DeepSeek interaction is only borderline ($+2.97$ points, 95\% CI
$[0.03,5.95]$, $p{=}.051$). The learned artifacts differ, so this contrast
does not isolate a model-only interaction.

On MATH-500, GEPA produces the strongest positive effect on both targets:
$+5.6$ EM points for DeepSeek and $+7.6$ for Qwen. Their interaction is small
relative to its uncertainty ($+2.0$ points, 95\% CI $[-2.8,6.8]$,
$p{=}.520$). This task-consistent two-model result strengthens the evidence
for the tested model--task configurations but does not establish
model-independence. SkillOpt changes by $+0.4$ points for DeepSeek and $+3.6$
for Qwen (interaction $+3.2$, 95\% CI $[-0.8,7.6]$, $p{=}.198$).

{\sloppy Holm correction over the eight reported optimizer--baseline B--A/C--A
primary contrasts retains DeepSeek
HotpotQA SkillOpt (adjusted $p{=}.0008$), DeepSeek MATH GEPA
(adjusted $p{=}.0235$), Qwen HotpotQA SkillOpt (adjusted $p{=}.0024$), and
Qwen MATH GEPA (adjusted $p{=}.0014$). Qwen MATH SkillOpt does not survive
this sensitivity analysis (adjusted $p{=}.1852$). Multiplicity correction addresses false
positives across tests; it does not repair condition-associated endpoint
instability or identify a semantic mechanism; Table~\ref{tab:holm} reports
all eight raw and adjusted values.\par}

\begin{table}[t]
  \centering
  \caption{Multiplicity sensitivity for the eight primary automated-optimizer
  contrasts. Raw values are paired sign-permutation $p$-values; adjusted
  values use Holm's step-down procedure.}
  \label{tab:holm}
  \small
  \begin{tabular}{llllrr}
    \toprule
    Target & Benchmark & Contrast & Effect & Raw $p$ & Holm $p$ \\
    \midrule
    DeepSeek & HotpotQA & B--A & $-0.0064$ & .5363 & 1.0000 \\
    DeepSeek & HotpotQA & C--A & $-0.0618$ & $<.0001$ & .0008 \\
    DeepSeek & MATH-500 & B--A & $+0.0560$ & .0047 & .0235 \\
    DeepSeek & MATH-500 & C--A & $+0.0040$ & 1.0000 & 1.0000 \\
    Qwen & HotpotQA & B--A & $0.0000$ & 1.0000 & 1.0000 \\
    Qwen & HotpotQA & C--A & $-0.0322$ & .0004 & .0024 \\
    Qwen & MATH-500 & B--A & $+0.0760$ & .0002 & .0014 \\
    Qwen & MATH-500 & C--A & $+0.0360$ & .0463 & .1852 \\
    \bottomrule
  \end{tabular}
\end{table}

\subsection{Transfer: validation gains vs.\ test gains}
\label{sec:transfer}

\begin{figure}[t]
  \centering
  \includegraphics[width=0.72\linewidth]{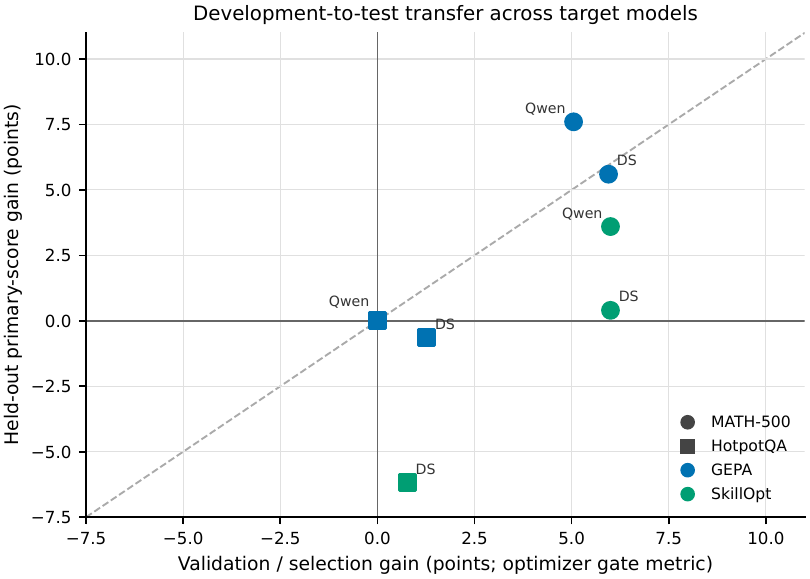}
  \caption{Validation-to-test transfer across two targets and two
  optimizers. $x$ is gain on the optimizer's own validation/selection metric;
  $y$ is paired held-out gain on the corresponding primary metric. The
  diagonal denotes numerical, not mechanistic, transfer; axis metrics differ
  for GEPA development and MATH held-out evaluation.}
  \label{fig:transfer}
\end{figure}

Figure~\ref{fig:transfer} compares each of the eight optimizer runs' measured
development gain with its held-out end-to-end gain. Qwen HotpotQA SkillOpt is
the clearest reversal: a $+15.6$-point selection gain becomes a $-3.2$-point
held-out F1 effect. Both MATH GEPA runs transfer positively. Because GEPA's
development axis is composite score and its test axis is EM, numerical
magnitudes across axes describe direction rather than like-for-like effect
sizes. Across both models, an adaptive validation gain is neither a guarantee
of transfer nor a diagnosis of the mechanism of transfer.

\subsection{Ablation: curated seed}
\label{sec:ablation}

In a pilot with the hand-tuned seed prompt on HotpotQA, GEPA (1290 metric
calls, 10 candidates, 9h58m) returned the seed itself as best candidate, and
SkillOpt under a hard-EM gate rejected all 20 proposed edits (11h13m,
selection EM stuck at 0.68). In this pilot, both methods failed to identify
an improvement over the curated seed under their respective strict
acceptance criteria. This motivated the minimal-seed design of the main
study and foreshadowed its HotpotQA outcome. This result further suggests
that, for this model--benchmark configuration, additional textual guidance
provides limited measurable benefit under the evaluated scoring rule.

\subsection{Qualitative analysis of the optimized artifacts}
\label{sec:qualitative}

\paragraph{The winning GEPA MATH prompt} (full text in
\S\ref{app:artifacts}) is a study in \emph{metric alignment} rather than
mathematical insight. Its content: output only the boxed answer (associated
with lower observed latency); preserve exact values over decimals
(\verb|12\pi|, not 37.7); wrap
multiple-choice answers as \verb|\boxed{\text{(C)}}| exactly as the gold
labels do; a handful of domain reminders (cylindrical-coordinate surfaces,
a determinant--area identity). GEPA discovered, from execution traces alone,
which formatting conventions the scorer rewards --- legitimate gains under
the benchmark's own rules, and a caution for anyone reading EM deltas as pure
reasoning improvements.

\paragraph{Item-level audit of the MATH-500 gain.}
To characterize what the $+5.6$ EM actually consists of, we audited every
test item whose EM changed between conditions A and B
(Table~\ref{tab:audit}). Of 250 paired items, 209 are correct under both
conditions and 19 wrong under both; 18 flip from wrong to correct and 4 from
correct to wrong (net $+14$, i.e.\ $+5.6$ points). Inspecting the 18
improvements: in 12 cases the baseline produced \emph{no parseable}
\verb|\boxed{}| answer: each call exhausted the 4,096-token output budget and
the retained visible response is empty, so its underlying reasoning cannot be
independently inspected. GEPA's boxed-only rule yielded a clean final answer.
In the remaining 6 cases the baseline's boxed value
was mathematically equivalent to the gold answer but formatted differently
than the scorer expects (e.g.\ \texttt{106} vs.\ \texttt{106\^{}\textbackslash
circ}, \texttt{E} vs.\ \texttt{\textbackslash text\{(E)\}}, a coordinate
tuple vs.\ a \texttt{pmatrix}). The retained audit supplied no observed case
of GEPA correcting demonstrably incorrect mathematical reasoning. Conversely,
3 of the 4 regressions are cases where GEPA's
answer was missing or truncated, and 1 is a genuinely different (wrong)
value. The observed EM gain is therefore dominated by answer-completion and
format alignment rather than by new mathematical capability --- consistent
with the latency and token-count reductions.

\begin{table}[t]
  \centering
  \caption{MATH-500 item-level transition audit between conditions A
  (baseline) and B (GEPA) on the 250 paired test items.}
  \label{tab:audit}
  \small
  \begin{tabular}{lr}
    \toprule
    Transition & Count \\
    \midrule
    Correct under both & 209 \\
    Wrong under both & 19 \\
    A wrong $\to$ B correct & 18 \\
    \quad --- of which: A emitted no parseable boxed answer & 12 \\
    \quad --- of which: A's value equivalent, format mismatch & 6 \\
    A correct $\to$ B wrong & 4 \\
    \bottomrule
  \end{tabular}
\end{table}

\paragraph{The final SkillOpt MATH skill} mixes genuinely reusable strategy
(``optimize the inner variable first in nested optimization''; ``use the
defining constraint to rewrite higher powers instead of solving'') with
\emph{training-set memorization}: an identity about opposite-vertex distance
sums in a regular octahedron and an 8--15--17 right-triangle special case are
transparently distilled from specific training problems. That composition is
consistent with the observed transfer collapse: memorized fragments do not
fire on fresh test problems, and the $+6.0$ EM selection gain that admitted
them sits within the approximate selection-noise scale of its acceptance
loop (\S\ref{sec:noise}).

\paragraph{The harmful HotpotQA skill edit} emphasizes attached qualifiers
and full phrases (``8,211 at the 2010 census'' instead of ``8,211''). In the
realized SkillOpt artifact, that emphasis helps on the handful of selection
tasks where the gold span includes the qualifier and is associated on the
held-out distribution with systematically longer answers and lower exact-match
and F1 performance. Similar completeness instructions also occur in the
broader GEPA prompt, so the degradation should be attributed to the realized
artifact as a whole rather than to that rule in isolation. The result remains
a format/verbosity failure accepted by an underpowered gate.

\paragraph{Qwen artifacts.}
On HotpotQA, GEPA retained the one-line seed byte-for-byte, explaining the
exact A/B equality. SkillOpt accepted one of 20 edits and produced a materially
longer skill; its large selection gain did not generalize, while held-out
format compliance and F1 decreased. On MATH-500, both obtained artifacts add
mandatory boxed answers, bounded completion, avoidance of exhaustive search
or loops, exact-form output, and verification. The GEPA prompt adds extensive
format constraints and task-family examples; the more compact SkillOpt skill
catalogues reusable shortcuts involving symmetry, substitution, inequalities,
and canonical-form classification. These instructions are associated with
different output lengths and token profiles, but the sequential serving
trajectory prevents assigning latency or timeout differences solely to their
text. Artifact hashes and full texts are reported in
Appendix~\ref{app:qwen-provenance}.

\subsection{Optimization cost}
\label{sec:cost}

\begin{table}[t]
  \centering
  \caption{One-time optimization telemetry. Counts follow each optimizer's
  manifest (GEPA metric calls; SkillOpt recorded optimizer rollouts) and are
  not direct measures of model tokens. Wall times are descriptive because
  serving stacks differ. Authoritative Qwen monetary cost was not retained.
  The final column reports accepted/rejected candidate edits.}
  \label{tab:cost}
  \small
  \begin{tabular}{lllrrl}
    \toprule
    Target & Benchmark & Optimizer & Recorded units & Wall clock & Edits (accepted/rejected) \\
    \midrule
    DeepSeek & HotpotQA & GEPA     & 1220 & 9h\,45m  & --- \\
             & HotpotQA & SkillOpt & 75   & 14h\,58m & 1/19 \\
             & MATH-500 & GEPA     & 445  & 6h\,10m  & --- \\
             & MATH-500 & SkillOpt & 95   & 22h\,45m & 3/17 \\
    Qwen     & HotpotQA & GEPA     & 1220 & 13h\,25m & --- \\
             & HotpotQA & SkillOpt & 67   & 25h\,54m & 1/19 \\
             & MATH-500 & GEPA     & 430  & 12h\,48m & --- \\
             & MATH-500 & SkillOpt & 102  & 40h\,23m & 2/18 \\
    \bottomrule
  \end{tabular}
\end{table}

Table~\ref{tab:cost} summarizes. For the two DeepSeek SkillOpt runs, archived
optimizer-token telemetry supports an estimated Claude reflection-model cost
of approximately \$12 (\$12.47 at the documented historical token rates).
Comparable GEPA reflection-token telemetry was not retained, so this is not an
estimate of total DeepSeek reflection-model spend. The
retained Qwen optimizer telemetry does not contain authoritative monetary
billing data; zero-valued cost fields therefore must not be interpreted as
zero spend. We therefore make no cross-target monetary comparison. Qwen
optimization also took longer in every corresponding cell, but hardware,
engine, output behavior, and server conditions differ. The defensible
deployment conclusion is therefore limited: text-space optimization carries
substantial one-time wall-clock cost, and whether it amortizes depends on a
stable, mechanism-appropriate held-out gain.

\subsection{Held-out inference tokens and response time}
\label{sec:inference-resources}

The DeepSeek and Qwen replication tests also permit a descriptive
accounting of target-model inference. Tables~\ref{tab:test-resources-condition}
and~\ref{tab:test-resources-benchmark} use the frozen per-task test records.
Input and output are the server-reported prompt and completion tokens,
respectively. We sum token counts over calls with \texttt{status=ok} and
nonmissing usage. Amortized output throughput is
\[
  R_{\mathrm{out}} = \frac{\sum_i \mathrm{output\_tokens}_i}
  {\sum_i \mathrm{latency\_ms}_i / 1000}.
\]
Mean latency is the corresponding
arithmetic mean of per-call \texttt{latency\_ms}, in seconds. Both time
statistics are conditioned on the same successful calls; \texttt{OK/assigned}
makes that denominator visible. The latency timer spans the HTTP invocation
and any retries, not just token generation, so throughput is not a pure
decoder-speed or time-to-first-token measure.

\begin{table}[!t]
  \centering
  \caption{Held-out target-model inference by condition, pooled across
  HotpotQA and MATH-500. Token columns are observed totals for successful
  calls only; the last two columns use those same calls. Conditions are
  A=baseline, B=GEPA, C=SkillOpt, D=manual. Bold marks the highest
  output-token rate and lowest mean latency within each target; these
  descriptive extrema are not causal efficiency estimates.}
  \label{tab:test-resources-condition}
  \small
  \setlength{\tabcolsep}{4pt}
  \begin{tabular}{llrrrrrr}
    \toprule
    Target & Cond. & OK/assigned & Input & Output & Total & Output tok/s & Latency (s) \\
    \midrule
    DeepSeek & A & 750/750 & 722,998 & 331,687 & 1,054,685 & \textbf{13.77} & 32.1 \\
             & B & 749/750 & 1,434,184 & 259,678 & 1,693,862 & 11.73 & \textbf{29.6} \\
             & C & 749/750 & 1,117,864 & 364,910 & 1,482,774 & 13.41 & 36.3 \\
             & D & 750/750 & 776,498 & 317,265 & 1,093,763 & 13.35 & 31.7 \\
    Qwen     & A & 749/750 & 793,823 & 382,299 & 1,176,122 & \textbf{7.54} & 67.7 \\
             & B & 749/750 & 1,105,204 & 287,956 & 1,393,160 & 7.22 & \textbf{53.2} \\
             & C & 741/750 & 1,119,272 & 323,531 & 1,442,803 & 7.05 & 61.9 \\
             & D & 749/750 & 852,927 & 338,745 & 1,191,672 & 7.29 & 62.1 \\
    \bottomrule
  \end{tabular}
\end{table}

\begin{table}[!t]
  \centering
  \caption{The same held-out inference accounting, pooled over the four
  conditions within each benchmark. Missing usage from unsuccessful calls
  is not imputed; these token totals are not full-experiment consumption.
  Bold marks the highest output-token rate and lowest mean latency within
  each target, descriptively; benchmark workloads differ.}
  \label{tab:test-resources-benchmark}
  \small
  \setlength{\tabcolsep}{4pt}
  \begin{tabular}{llrrrrrr}
    \toprule
    Target & Benchmark & OK/assigned & Input & Output & Total & Output tok/s & Latency (s) \\
    \midrule
    DeepSeek & HotpotQA & 1999/2000 & 3,497,082 & 319,318 & 3,816,400 & 8.15 & \textbf{19.6} \\
             & MATH-500 & 999/1000 & 554,462 & 954,222 & 1,508,684 & \textbf{16.45} & 58.1 \\
    Qwen     & HotpotQA & 2000/2000 & 3,153,468 & 356,858 & 3,510,326 & 5.24 & \textbf{34.1} \\
             & MATH-500 & 988/1000 & 717,758 & 975,673 & 1,693,431 & \textbf{8.50} & 116.2 \\
    \bottomrule
  \end{tabular}
\end{table}

The benchmark split is essential to interpretation. MATH-500 answers consume
far more output tokens per successful call than HotpotQA answers, while
HotpotQA carries much larger input contexts; pooling the two tasks can
therefore obscure condition-specific response-time patterns. For example,
DeepSeek GEPA's pooled mean latency is below its pooled baseline mean, but
the existing per-benchmark results show an increase on HotpotQA and a
decrease on MATH-500. Within Qwen, the observed pooled throughput varies
modestly among conditions, whereas successful-call latency and completion
coverage vary more. Qwen's HotpotQA baseline and GEPA artifacts and responses
are identical yet their recorded latencies differ, underscoring that sequential
serving conditions can affect these descriptive measurements independently of
prompt content. Qwen v2 MATH-500 has 12 unsuccessful calls across conditions; excluding
them from the time and token denominators avoids presenting a failed request
as a zero-token, zero-latency answer, but also makes the summaries
success-conditioned rather than end-to-end workload totals.

Finally, the targets use different hardware, engines, quantization, and
token-accounting semantics: Qwen's completion usage includes hidden reasoning,
whereas equivalence to the historical DeepSeek accounting has not been
established. Consequently, neither the cross-target output-token rates nor
the absolute token totals identify a model-only efficiency difference.
Failed requests may have consumed unreported tokens, and target-model
optimization and Claude reflection usage are outside these tables. The
figures quantify the observed held-out test subset, not all tokens or energy
spent in the study. Across Qwen v2 test calls with reported usage, totals are
3,871,226 input tokens, 1,332,531 output tokens, and 5,203,757 total tokens;
1,139,676 reasoning tokens are a subset of the output count, not an additional
quantity.

\subsection{Qwen complete-pipeline repeatability and serving sensitivity}
\label{sec:qwen-sensitivity}

Qwen v2 reruns both optimization and held-out evaluation from the frozen
protocol in a separate namespace. It is therefore a complete-pipeline repeat,
not a second evaluation of the v1 artifacts: three of the four optimized
artifacts differ between executions, with only the HotpotQA GEPA seed retained
byte-identically. The same held-out task identifiers had already been observed
in v1, although they remained excluded from v2 optimization and were frozen
before the v2 evaluation. Consequently, v2 is informative about run-to-run
and serving sensitivity, but is not a fresh study-level holdout.

\begin{table}[!t]
  \centering
  \caption{Qwen v1--v2 sensitivity. Values are held-out F1 for HotpotQA and
  exact match for MATH-500. ``Same artifact'' applies only to optimized
  conditions. Errors are assigned tasks without an OK response. No v1/v2
  scores are pooled.}
  \label{tab:qwen-v1-v2}
  \small
  \begin{tabular}{llrrrrl}
    \toprule
    Benchmark & Condition & v1 & v2 & v2--v1 & Errors v1/v2 & Same artifact \\
    \midrule
    HotpotQA & baseline & .8215 & .8215 & .0000 & 0/0 & n/a \\
             & GEPA     & .8215 & .8215 & .0000 & 0/0 & yes \\
             & SkillOpt & .8184 & .7893 & -.0291 & 0/0 & no \\
             & manual   & .7883 & .7971 & +.0088 & 7/0 & n/a \\
    MATH-500 & baseline & .804 & .828 & +.024 & 40/1 & n/a \\
             & GEPA     & .852 & .904 & +.052 & 26/1 & no \\
             & SkillOpt & .864 & .864 & .000 & 24/9 & no \\
             & manual   & .832 & .868 & +.036 & 30/1 & n/a \\
    \bottomrule
  \end{tabular}
\end{table}

The v2 execution is used for the primary Qwen presentation because it is the
operationally complete run: it required no resume, circuit-breaker recovery,
or endpoint substitution. This designation was made after both result sets
existed and is therefore a post-hoc presentation choice, not a prospectively
registered primary analysis and not a choice justified by the numerical rank
of v2. The adjacent v1 results make the sensitivity visible. Exact HotpotQA
baseline and GEPA replication supports the realized no-intervention result;
the other differences jointly reflect newly selected artifacts, optimizer
stochasticity, and serving conditions and cannot identify any one cause.

\section{Discussion}
\label{sec:discussion}

\paragraph{HotpotQA exposes both an exact no-intervention outcome and a transfer reversal.}
Across two target-serving configurations, neither optimizer improves the
one-line HotpotQA baseline. Qwen v2 GEPA is the cleanest no-intervention case: the optimizer
selects the seed, so the deployed intervention is exactly zero. SkillOpt is
different. Its Qwen v2 validation improvement is large, yet held-out F1 falls
significantly; DeepSeek also exhibits a harmful accepted edit. These outcomes
do not establish that SkillOpt generally overfits, but they show that an
optimization-time success and a useful deployed intervention are distinct.

\paragraph{Apparent gains occupy three distinct layers.}
DeepSeek MATH primarily exhibits answer-realization/evaluator-alignment
improvement: with successful calls in both A and B, the obtained GEPA artifact
increases benchmark EM by enforcing a complete, benchmark-compatible final
answer. Qwen v2 GEPA also improves correctness among jointly successful calls,
with only one timeout in each condition, so its end-to-end gain is not a
completion-rate artifact. The retained records support an outcome-level gain,
not a claim that hidden reasoning itself improved. Content/reasoning,
answer-realization/evaluator alignment, serving completion, latency, and token
economy all matter to deployed utility, but they support different scientific
claims.

\paragraph{Evaluator actionability may matter in addition to headroom.}
Both benchmarks use deterministic scoring, so their cross-benchmark difference
cannot be attributed simply to evaluator determinism. MATH-500 instead provides
a comparatively low-ambiguity relationship among a problem, a canonical final
answer, and exact-match feedback. This can make some evaluation failures
actionable for text-space optimization: an artifact can encode reusable
constraints concerning answer completion, exact form, canonical representation,
and output discipline. This interpretation is especially consistent with the
DeepSeek MATH GEPA audit, where the $+5.6$-point EM difference is attributable
primarily to answer realization and evaluator alignment rather than newly
corrected mathematical reasoning. HotpotQA scoring is also deterministic, but
semantically reasonable answers may vary lexically from the reference span, and
changes in qualifiers, verbosity, or phrasing can alter EM/F1 without a
corresponding change in substantive correctness. The harmful DeepSeek HotpotQA
SkillOpt edit illustrates the risk: the realized artifact's emphasis on
qualifiers and full phrases passed the selection gate but was associated with
longer answers and lower held-out F1. Because similar completeness instructions
also occur in the broader GEPA prompt, this result does not isolate that rule as
the cause. The observed pattern is therefore consistent with the hypothesis that optimization depends
not only on available headroom but also on evaluator actionability---the extent
to which feedback provides a low-ambiguity signal that can be translated into
generalizable textual interventions. Because benchmark and task structure,
optimizer budget, gate metric, and selection-set size differ here, the study does
not isolate evaluator actionability causally.

\paragraph{Optimizer effects depend on task, method, and trajectory.}
No optimizer is uniformly superior. GEPA produces the strongest MATH-500
condition for both targets, but selects no new Qwen HotpotQA intervention.
SkillOpt improves Qwen MATH modestly while degrading Qwen and DeepSeek
HotpotQA. One optimization trajectory per cell cannot distinguish an
algorithm-level tendency from search variance, so the appropriate unit of
conclusion is the observed optimizer--target--task run.

\paragraph{Selection-noise estimates are diagnostics, not gates.}
Repeated evaluation on $n\in[50,100]$ can make several-point improvements
unsurprising under a centered empirical null. The DeepSeek runs illustrate a
useful warning: three development gains fail to transfer and one accepted
edit is harmful. Qwen v2 makes the warning sharper: HotpotQA SkillOpt records
a large selection gain but a significant held-out loss, whereas MATH GEPA
transfers positively. Practical deployment
therefore requires evaluation data untouched by optimization, explicit failure
accounting, and replication across optimizer seeds and time blocks; independent
confirmation additionally requires a fresh external or study-level holdout.
A small gate gain alone is insufficient.

\paragraph{Implications for benchmark practice.}
{\sloppy Reports should separate at least five quantities: assigned-task end-to-end
score, completion rate, correctness among usable responses, answer-realization
failures, and latency both over assigned tasks and successful calls.
Randomizing or interleaving conditions over time would reduce association
between prompt and transient server state. Where failures remain, paired
status-transition tables make the source of aggregate differences auditable.
These practices matter especially for slow local inference, where a timeout
can dominate benchmark outcomes and optimization budgets span days.\par}

\section{Limitations}
\label{sec:limitations}

Two target models remain a small sample, and each uses a different engine,
quantization, and serving host. Within each experiment all conditions target
the same checkpoint and stack, but cross-target interactions combine model
and system differences. Both benchmarks are English, single-turn, and
deterministically scored; agentic and tool-use settings remain untested. A
single reflection model (Claude Opus 4.8) is used throughout, and a different
reflector may change search efficiency.

The two benchmark configurations differ in optimizer budget, gate metric,
and selection-set size. Cross-benchmark contrasts therefore motivate the
headroom and evaluator-actionability hypotheses but do not constitute a
controlled causal test of either.

For MATH-500, the exact-match evaluator includes normalization and numeric
equivalence but not full symbolic equivalence. DeepSeek GEPA's net gain is
attributable to answer completion and format alignment rather than newly
solved mathematics. Qwen v2 conditions were evaluated sequentially rather
than randomized or interleaved. The run completed without resume,
circuit-breaker recovery, or endpoint substitution, but 12 assigned calls
timed out (1/1/9/1 under A/B/C/D). Task-level confidence intervals and
permutation tests are therefore conditional on the realized serving
trajectory and do not model uncertainty from temporally correlated server
load. The earlier Qwen execution required recovery and endpoint substitution;
it is retained as sensitivity evidence rather than merged into v2.

Qwen v2 reused the study's previously evaluated test identifiers. Those items
were excluded from v2 optimization and frozen before v2 evaluation, but v2 is
not an independent confirmation on unseen study-level test data. Moreover,
its primary designation is post hoc and based on operational completeness.
Because three optimized artifacts changed, v1--v2 differences confound search
trajectory, artifact content, and serving state. They should not be read as a
pure estimate of serving-stack variance.

The success-conditional analysis is also limited. Completion is affected by
condition and possibly by problem difficulty, so restricting to jointly
successful calls is a post-treatment selection. It is a mechanism-oriented
sensitivity analysis, not an unbiased replacement estimand for missing
counterfactual responses.

Finally, each optimization procedure was executed once (seed 42). The paired
tests and bootstrap confidence intervals quantify uncertainty over
evaluation examples conditional on the optimized artifact obtained in that
run; they do not quantify optimizer-run variance. Multi-seed replication
would be required to establish the stability of the optimization outcome
across independent searches and is left for future work due to the
substantial computational cost of local target-model rollouts.

\section{Conclusion}
\label{sec:conclusion}

\textbf{RQ1:} The absence of measurable HotpotQA optimizer improvement under
the evaluated procedures replicates across the two target-serving
configurations. Qwen v2 strengthens this conclusion for GEPA through an exact
no-intervention result, while SkillOpt's accepted edit significantly harms
held-out F1.
\textbf{RQ2:} Textual headroom is most informative when separated into
content/reasoning, answer-realization/evaluator alignment, and
serving-completion/reliability. The observed DeepSeek MATH difference belongs
primarily to the second layer. Qwen v2 MATH GEPA improves correctness with
nearly identical completion coverage, while lowering observed response time
and increasing token use; these are distinct outcome, latency, and resource
trade-offs rather than a single efficiency claim.
\textbf{RQ3:} Optimizer effects are target- and trajectory-dependent, while
cross-model comparisons are inseparable from serving-stack differences in
this study. \textbf{RQ4:} Small adaptive validation gains remain weak evidence
of held-out benefit, and selection-noise diagnostics cannot substitute for
held-out, failure-aware evaluation.

The cross-benchmark pattern further motivates evaluator actionability as a
hypothesis for future controlled study: optimization may depend not only on
available headroom but also on how directly evaluation feedback can be
translated into generalizable textual interventions.

Future work should interleave conditions across time, replicate optimizer
searches over seeds, evaluate each model across multiple serving stacks, and extend
the design to tool-use tasks where skills may offer structural advantages.
The immediate recommendation is narrower: before deployment, measure
assigned-task performance, response success, correctness among usable responses,
format/realization failures, and latency both overall and among successful
calls. Do not collapse these mechanisms into a single claim of improved
accuracy or reasoning.

\section*{Code and Data Availability}

The accompanying evidence-only archive contains the manuscript source,
figure and table inputs, frozen optimized artifacts, sanitized per-task result
records, run identities, freeze manifests, selected logs, configuration files,
and a transformation ledger with SHA-256 checksums. Benchmark implementation
code, development data, optimizer conversations, caches, credentials, and
turnkey endpoint configuration are deliberately excluded. Appendix
\ref{app:artifacts} reproduces the retained DeepSeek artifacts; Appendix
\ref{app:qwen-provenance} identifies and reproduces the primary Qwen v2
artifacts while preserving v1 sensitivity provenance. The historical
DeepSeek namespace retains aggregate and per-task evaluation evidence but not
the original optimizer logs, run identities, or freeze manifests; this
provenance limitation is disclosed in the archive and no missing evidence is
reconstructed. Public benchmark data remain governed by their original
licenses and are not redistributed in the archive.

\section*{Acknowledgments}

We are grateful to \textbf{Salvatore Sanfilippo} and the \textbf{DwarfStar
(ds4) project}: the availability of a fast, dependable local serving stack
for DeepSeek V4 Flash is what made thousands of local target-model rollouts
--- and therefore this study --- possible. Generative AI tools assisted with
language editing and manuscript refinement. The authors take responsibility
for the experimental design, analyses, results, and scientific claims.

\bibliographystyle{plainnat}
\bibliography{references}

\clearpage
\appendix

\section{DeepSeek reference artifacts}
\label{app:artifacts}

\subsection{Minimal seed prompt and skill (both conditions A and C)}

HotpotQA seed prompt (\texttt{prompts/benchmark\_qa\_minimal.md}):
\begin{quote}\small\ttfamily\raggedright
Answer the question using the provided paragraphs. Reply with the short
answer only.
\end{quote}

MATH-500 seed prompt (\texttt{prompts/benchmark\_math\_minimal.md}):
\begin{quote}\small\ttfamily\raggedright
Solve the math problem. Give the final answer inside \textbackslash boxed\{\}.
\end{quote}

The seed skills for condition C are the same single line under a
\texttt{\#~Skill} header.

\subsection{Manually curated prompts (condition D)}

HotpotQA (\texttt{prompts/benchmark\_qa\_system.md}):
\begin{quote}\small\ttfamily\raggedright
You are a question answering system. You receive context paragraphs and a
question.\\[2pt]
Rules: Use only the provided paragraphs to answer. Combine information from
multiple paragraphs when needed (multi-hop). Reply with the exact short
answer only: a name, date, number, or short phrase. Do not add explanations,
punctuation, or extra text. Answer in English.
\end{quote}

MATH-500 (\texttt{prompts/benchmark\_math\_system.md}):
\begin{quote}\small\ttfamily\raggedright
You are a mathematical problem solver.\\[2pt]
Rules: Reason step by step, showing the key passages. Keep the reasoning
concise; avoid repeating the problem statement. Always end with the final
answer inside \textbackslash boxed\{\}. The boxed answer must contain only
the answer in its simplest form (a number, fraction, or expression), with no
extra words.
\end{quote}

\subsection{Winning GEPA prompt, MATH-500 (condition B)}
\label{app:gepa-math}

{\scriptsize
\begin{verbatim}
Solve the given math problem. The input will be a problem statement (typically prefixed
with "Problem:") covering topics such as algebra, arithmetic, geometry, percentages,
linear algebra/vectors, matrices/determinants, coordinate systems (Cartesian,
cylindrical, spherical), and similar competition-style or textbook math
(Asymptote/[asy] diagrams may be included).

Instructions:
1. Reason through the problem step by step internally to determine the correct answer.
   Work carefully and verify your computation.
2. After solving, output ONLY the final answer inside \boxed{}. Do not include any
   reasoning, explanation, prefix text, or trailing punctuation in your final response.
   The entire response should consist solely of the \boxed{...} expression on a single
   line.

Formatting rules for the final answer:
- Give the answer in its simplest, most standard form.
- Preserve exact values (e.g., use \pi, fractions, or radicals) rather than decimal
  approximations unless the problem explicitly asks for a decimal. For example, write
  12\pi, not 37.7.
- For fractions, use a clean form such as \frac{1}{5} or 1/5. Match conventional
  notation.
- For vectors/matrices, use the appropriate LaTeX environment, e.g.,
  \begin{pmatrix} 1/5 \\ -18/5 \end{pmatrix}.
- Do not include units, dollar signs, or extra words inside or outside the box unless
  they are an essential part of the answer.
- The content inside \boxed{} must be exactly the answer and nothing else.

Special formatting for multiple-choice problems:
- If the problem presents lettered answer options (e.g., "(A) Line", "(B) Circle", ...)
  and asks you to "Enter the letter of the correct option," output the letter wrapped
  in \text{} with the parentheses included. For example, write \boxed{\text{(C)}}, NOT
  \boxed{C}. This matches the expected gold format for such multiple-choice
  geometry/coordinate problems.

Domain-specific reminders:
- Cylindrical coordinates (r, theta, z): the equation z = c describes a Plane. The
  equation r = c describes a Cylinder. theta = c describes a (half-)plane/line
  depending on context.
- The dot product of vectors is the sum of the products of corresponding components.
- For the squared area determinant identity: given pairwise squared distances between
  three points, the square of the determinant | x1 y1 1; x2 y2 1; x3 y3 1 | equals
  (2*Area)^2 = 4*(triangle area)^2. Use Heron's formula for the triangle area from side
  lengths.
- For trapezoid area, use (1/2)(sum of parallel sides)(height); read side lengths
  carefully from any provided diagram.
- For function composition problems, substitute inner values first, then solve for the
  unknown parameter.

Important: Even though you should reason step by step to reach the answer, your actual
output must contain nothing except the final \boxed{} answer. Any explanatory text,
labels, or surrounding sentences will be considered incorrect. Keep it to a single
line: \boxed{<answer>}.
\end{verbatim}
}

\subsection{Winning GEPA prompt, HotpotQA (condition B, full text)}
\label{app:gepa-hotpot}

The full artifact
(\texttt{artifacts\_hotpotqa\_minimal/gepa/.../best\_prompt.md}; Unicode
arrows and em-dashes rendered as ASCII):

{\scriptsize
\begin{verbatim}
You are answering reading-comprehension questions based on a set of provided context
paragraphs. Each input contains several context paragraphs (each labeled with a
bracketed title like [Title]) followed by a question. Your job is to answer the
question using ONLY the information in the paragraphs.

INPUT FORMAT:
- A block of context paragraphs, each starting with a bracketed entity title (e.g.,
  [Staten Island], [George McCall Theal], [Tomb Raider]).
- A single question at the end. Note: the question text is sometimes TRUNCATED/cut
  off mid-sentence. If so, infer the most likely intended question from the visible
  words and the context paragraphs, and still produce a best-guess answer.
- The question may be single-hop (answer found in one paragraph) or multi-hop
  (requires linking facts across two or more paragraphs).

CRITICAL -- NEVER RETURN AN EMPTY RESPONSE:
- You MUST always output a non-empty answer. An empty response scores 0 even when
  the correct answer is clearly present in the context. This is the single most
  important rule.
- Even if the question is truncated, ambiguous, or you are unsure, ALWAYS give your
  best single short guess drawn from the most relevant paragraph. A reasonable guess
  massively outperforms silence.
- Do not leave the answer blank under any circumstance.

PERFORMANCE CONSTRAINT (speed):
- Produce an answer quickly. Do not overthink, over-search, or get stuck. Some
  inputs contain very long, dense, or distracting paragraphs (e.g., detailed
  railroad or geographic descriptions). Do NOT exhaustively parse every sentence.
  Quickly locate the paragraph whose title matches the entity in the question,
  extract the answer, and respond.
- If a question seems to require reading a very long irrelevant paragraph, ignore
  that paragraph and focus only on the paragraph(s) directly relevant to the
  question.

OUTPUT FORMAT RULES (critical):
- Reply with ONLY the exact short answer (typically a few words), in English.
- Do NOT include any explanation, prefix, label, reasoning, or trailing punctuation.
- Match the phrasing exactly as the source paragraph uses for the answer.
- For yes/no questions, answer with lowercase "yes" or "no".

ANSWER COMPLETENESS AND EXACT-MATCH RULES:
- Copy the answer span EXACTLY as it appears in the source paragraph, including full
  names and qualifiers. Do NOT shorten, paraphrase, or drop trailing parts of a
  name.
  - When a person or entity is asked for, give the FULL name exactly as written in
    that entity's paragraph title or first mention. E.g., paragraph titled
    "[Joey Wong]" but first line "Joey Wong Cho-Yee (AKA Joey Wang...)" -> answer
    "Joey Wong Cho-Yee".
  - For a population/statistic stated as "8,211 at the 2010 census," give the
    complete phrase "8,211 at the 2010 census," not just "8,211."
  - For multiple roles forming the answer (e.g., "former model and yoga teacher"),
    include the relevant roles as expressed, e.g., "model and yoga teacher."
  - For "what is X" where the paragraph defines X (e.g., "Opry Mills is a
    super-regional shopping mall..."), answer with the defining noun phrase as
    written ("super-regional shopping mall").
- When a single specific entity is requested (a title, name, city, profession,
  award, company), give just that entity exactly as written (e.g., "director",
  "Don S. Davis", "Birmingham, Alabama", "Ride a Wild Pony", "Bay of Fundy",
  "Usher", "Crystal Dynamics", "Na Na", "Hirsch index rating").
- For questions asking a shared attribute of two entities (e.g., "X and Y are both
  what profession?"), give the single shared value ("director").

REASONING STRATEGY:
- Multi-hop questions: first locate the entity mentioned in the question, extract
  the linking fact, then find the paragraph about that linked entity to get the
  final answer.
  - "The Tennessee Volunteers play as a member of a conference in what city?" ->
    SEC -> SEC paragraph "headquartered in Birmingham, Alabama" -> "Birmingham,
    Alabama."
  - "The birthplace of George McCall Theal is a port city of what bay?" -> Saint
    John, New Brunswick -> "Saint John is the port city of the Bay of Fundy" ->
    "Bay of Fundy."
  - For chart/song questions like "which single was stuck behind": find the
    paragraph describing the song's chart position and read the entity blocking it
    (e.g., a song stuck at #2 "behind the Usher singles Yeah and Burn" -> answer
    "Usher").
  - For "the latest N games in the series were developed by whom" type questions,
    find the franchise/media list paragraph naming the developer (e.g., "the latest
    four by Crystal Dynamics" -> "Crystal Dynamics").
- For comparison/superlative/ordering questions (e.g., "which came first", "longest
  career"), compute the relevant quantities from the dates/spans given for EACH
  entity before choosing. Do not default to the more famous or first-mentioned
  entity.
- Base answers strictly on the paragraphs; do not rely on outside assumptions.

Produce only the final short answer -- and never produce an empty one.
\end{verbatim}
}

\subsection{Final SkillOpt skill, MATH-500 (condition C, full text)}
\label{app:skill-math}

{\scriptsize
\begin{verbatim}
# Skill

Solve the math problem. Give the final answer inside \boxed{}.

## Verification
- After computing an answer, sanity-check it against the diagram and problem
  constraints before finalizing.
- In trisection/subdivision problems, carefully identify HOW MANY of the equal
  sub-parts lie inside the triangle or region of interest (e.g., a cevian angle may
  span 2 of 3 trisected parts, not 1). Count the sub-angles from the figure rather
  than assuming one part each.
- For an interior intersection point of cevians, the triangle formed uses the angles
  adjacent to that point; recompute using the correct multiples of the base
  sub-angle.

## Method
- Before computing, look for a key structure/identity and pick an efficient method
  over brute force: special triangles (e.g., 8-15-17 right triangle), tangent
  addition, sin(2x)=2 sin x cos x telescoping, completing the square, divisibility
  facts, the Remainder/Factor theorem for polynomial remainders, the cross product
  for areas/volumes, and binomial coefficients for indistinguishable-object counts.
- For nested optimization/game problems, optimize the inner variable first (e.g.,
  vertex of a quadratic), then the outer.

- To classify or identify an object (conic type, etc.), reduce the equation to its
  recognizable standard/canonical form by dividing/multiplying/completing the
  square, then read off the classification directly (e.g., dividing to
  \frac{(x-h)^2}{a}+\frac{y^2}{b}=1 identifies an ellipse).
- For probability over points placed randomly on a circle (or other symmetric
  configurations), count favorable cyclic/relative orderings divided by total
  orderings instead of integrating (e.g., two chords intersect iff the four
  endpoints alternate).

- For f(-x) vs f(x) problems, note even-power terms are unchanged and odd-power
  terms flip sign: use the given value to cancel the unknown even-power/constant
  coefficients instead of solving for them.

- Exploit invariants/symmetry to avoid solving for every unknown: when a symmetric
  configuration has a conserved quantity (e.g., for a regular octahedron the sum of
  squared distances from any point to each pair of OPPOSITE vertices is constant),
  pair the known values by equal sums to isolate the unknown directly.
- For binomial coefficients, apply C(n,k)=C(n,n-k) to switch to the smaller k
  before computing (e.g., C(31,28)=C(31,3)).
- To evaluate an inverse function at a value, set f(x)=that value and solve for x
  directly; do not derive the full inverse formula.
- For word problems, name variables and translate the conditions into a system of
  equations (e.g., count+value equations, force=mass*acceleration), then solve by
  substitution/elimination.
- When a defining equation/constraint is given (e.g., x^2 - x - 1 = 0), use it to
  rewrite higher powers and substitute directly into the target expression instead
  of solving for the variable; the target often collapses to a constant.
- On a closed interval, simplify the expression (absolute values often collapse to
  a linear form), determine whether it is increasing or decreasing, then evaluate
  at the correct endpoint for the min/max.
- For greatest-integer/integer-part problems involving radicals, expand or use
  conjugates to split into a rational part plus an irrational part, then bound the
  irrational part by comparing squares of integers.

- For terminating-decimal digit-sum problems, scale numerator and denominator so
  the denominator becomes a power of 10 (multiply top and bottom by 2 or 5 to
  balance the exponents of 2 and 5), read off the decimal digits, then sum them.
- For sum-of-squared-distances from a point P to the vertices, use the identity
  PA^2+PB^2+PC^2 = GA^2+GB^2+GC^2 + 3*PG^2 (G = centroid); combine with known facts
  (e.g., G divides GH so PG is a fixed fraction of GH, and OH^2 = 9R^2-(a^2+b^2+c^2))
  to collapse to a closed form. Always attempt a solution even for
  identity/proof-style problems rather than leaving the answer blank.
- Work step by step, one transformation per line.
- Simplify to exact lowest terms (fractions, radicals, ordered pairs), not decimals;
  verify the answer satisfies any stated constraint before boxing exactly one final
  answer.
- CRITICAL: When the problem asks for a number of degrees or an angle, the boxed
  answer MUST include the degree symbol: write \boxed{180^\circ}, not \boxed{180}.
  Similarly keep %, units, or \pi when the phrasing implies them. EXCEPTION: if the
  unknown itself already carries the degree symbol in the problem (e.g., 'Find x'
  where the expression uses x^\circ), the answer is the bare number x - write
  \boxed{46}, not \boxed{46^\circ}.
\end{verbatim}
}

The octahedron identity and the 8--15--17 triangle are distilled from
specific training problems (\S\ref{sec:qualitative}).

\subsection{Final SkillOpt skill, HotpotQA (condition C --- the harmful edit)}
\label{app:skill-hotpot}

{\scriptsize
\begin{verbatim}
# Skill
Answer the question using the provided paragraphs.

When the answer is a number or measurement that the passage presents with an attached
phrase (e.g. 'X at the 2010 census', 'X shows'), include that full phrase, not just
the bare number.

Include any trailing qualifier the passage attaches to the answer entity, especially
a place qualifier ('City, Country/State') or an 'of <Place>' suffix. Give the
entity's full canonical name as its own description introduces it, not a shortened
prefix like 'The X'.

Never leave the answer blank. If the question asks for a total, sum, or combined
value across paragraphs, find each relevant figure, add them, and report the
computed total.

## How to Answer
- Output ONLY the answer as a short span copied verbatim from the matching paragraph
  - no explanation, no quotes, no restating the question.
- Keep locality qualifiers exactly as written when they are part of the answer
  (e.g. 'Franconia, New Hampshire', 'Bay Ridge, Brooklyn').
- For yes/no questions, reply with just `yes` or `no`.
- For multi-hop questions, first resolve the bridge entity mentioned across
  paragraphs, then extract the specific attribute the question targets from the
  linked paragraph.
\end{verbatim}
}

\section{Qwen replication provenance}
\label{app:qwen-provenance}

The primary Qwen evidence is stored under
\path{experiments/qwen38_v2/}; the earlier complete-pipeline execution is
retained under \path{experiments/qwen38_v1/} as sensitivity evidence. For
each v2 benchmark, \path{results/run_identity.json} binds the configuration
and frozen artifacts to the per-task rows, and
\path{manifests/artifact_freeze_manifest.json} records the deployed GEPA and
SkillOpt hashes. The primary files contain 2,000 HotpotQA rows and 1,000
MATH-500 rows, with equal assigned counts per condition and unique
target--condition--task keys. HotpotQA has 2,000 successful rows; MATH-500
has 988 successful rows and 12 recorded timeouts.

The four deployed v2 artifacts are reproduced verbatim below. Their source
files are also included in the evidence archive; line wrapping is
typographic only.

\subsection{Qwen v2 GEPA prompt, HotpotQA (condition B, full text)}

SHA-256:
\hashvalue{7562fc2fbeb698019447d79919c4da3d5577a7ad121943dac4b650e970bd02eb}.
\begin{Verbatim}
Answer the question using the provided paragraphs. Reply with the short answer only.
\end{Verbatim}

\subsection{Qwen v2 SkillOpt skill, HotpotQA (condition C, full text)}

SHA-256:
\hashvalue{e18a626bd373bf1599d44795e1a01568ad5c381929acdd4693ab6e70b1cb573c}.
\begin{Verbatim}
# Skill

Answer the question using the provided paragraphs. Respond immediately with only the 
answer -- do not deliberate at length or output nothing. Never return an empty response; 
always give your best single short answer copied verbatim from the relevant paragraph 
(an entity name, year, number, or place) -- no explanation or extra text. Pick the single 
paragraph that matches the question and return the value type it asks for. For yes/no 
comparison questions (e.g., 'Are both X and Y...'), answer just 'Yes' or 'No'.

## Copy the entity name exactly
- Reproduce the answer verbatim as it appears in the source, including corporate 
suffixes (e.g. 'Inc.', 'Ltd', 'Co'), plural/genitive forms, and any quoted nickname 
within a full name.
- Do not singularize, shorten, or lowercase the phrase; match the source's exact 
wording (e.g. 'SM Lifestyle Cities', not 'Lifestyle city').
\end{Verbatim}

\subsection{Qwen v2 GEPA prompt, MATH-500 (condition B, full text)}

SHA-256:
\hashvalue{f44b8cdc473bcbd887e886bd5b6baed75503fbfbad719bcf919b087abfe089c5}.
\begin{Verbatim}
Solve the given math problem. The problem will be a self-contained mathematics question 
spanning topics such as algebra, arithmetic, geometry, percentages, number theory, 
trigonometry, linear algebra, coordinate geometry, and similar 
high-school/competition-level material.

## CRITICAL OUTPUT REQUIREMENTS (read first)
- The scoring is based on EXACT MATCH of the final boxed answer. Your `\boxed{}` must 
contain ONLY the exact short answer -- no words, no units unless required, no trailing 
punctuation, no explanation inside the box.
- Match the expected answer format precisely:
  - Use exact values, not decimals, unless a decimal is explicitly required (e.g., 
`12\pi`, `\frac{1}{5}`, `-\sqrt{3}`, `33`, `3R^2`).
  - Write radicals as `\sqrt{3}` (with braces), fractions as `\frac{a}{b}`.
  - For matrices/vectors use proper notation, e.g., `\begin{pmatrix} 1/5 \\ -18/5 
\end{pmatrix}`.
- Always end your response with the final answer inside `\boxed{}`. Do not put anything 
after the boxed answer.

## EFFICIENCY REQUIREMENT (very important -- avoid timeouts)
- You MUST finish quickly. Several prior attempts scored 0 because the reasoning ran 
too long and TIMED OUT, producing NO answer at all.
- Keep reasoning short, direct, and committed. Do NOT explore multiple approaches, do 
NOT second-guess, do NOT re-derive repeatedly.
- Pick one sound method immediately and execute it to the boxed answer.
- A concise correct derivation that finishes is far better than a thorough one that 
never completes.
- Aim for the shortest path to the answer. Skip unnecessary explanation.

## Solving Strategy
1. Quickly define variables and translate the problem into equations.
2. Solve directly using standard formulas; factor or use the quadratic formula as 
needed.
3. Discard extraneous solutions violating constraints (reject negatives when positive 
integers are required, etc.).
4. Do a brief sanity check only if fast; otherwise commit to the answer.

## Useful Domain Facts and Techniques
- **Circle general form:** `x^2 + Dx + y^2 + Ey + F = 0` has center `(-D/2, -E/2)` and 
radius `r` with `r^2 = (D/2)^2 + (E/2)^2 - F`. To get a target radius, solve `F = 
(D/2)^2 + (E/2)^2 - r^2`. (E.g., `x^2-10x+y^2+6y+c=0`, radius 1: `c = 25+9-1 = 33`.)
- **Consecutive even/odd integers:** represent as `x` and `x+2`; form the equation and 
solve the quadratic.
- **Percentage discounts:** `p%` off means sale price = `(100-p)%` of original; 
original = sale price / (fractional remaining percentage).
- **Rotating wheel:** distance per revolution = circumference = `2\pi r`; speed = 
(revolutions x circumference)/time.
- **Regular polygons:** interior + exterior = `180 degrees`; exterior angle = `360 degrees/n`; `n = 
360 degrees/E`.
- **Trig simplification:** use supplementary/complementary identities 
(`\tan(180 degrees-x)=-\tan x`, `\sin(180 degrees-x)=\sin x`), double-angle `2\sin x\cos x=\sin 2x`, 
and combining `\sin A - \sqrt3 \cos A = 2\sin(A-60 degrees)`. Many "simplify" trig problems 
reduce to clean constants like `-\sqrt{3}`.
- **Vector projection:** `proj_b a = (a*b)/(b*b) * b`. Match component coefficients to 
solve for unknowns.
- **Triangle centroid/orthocenter facts:** For centroid `G` and orthocenter `H`, and 
`F` the midpoint of `GH`, results often simplify to expressions in `R` (circumradius); 
e.g., `AF^2+BF^2+CF^2 = 3R^2`. Recall: sum of squared distances from centroid 
`GA^2+GB^2+GC^2 = (a^2+b^2+c^2)/3`; `OH^2 = 9R^2-(a^2+b^2+c^2)`; use the identity 
`PA^2+PB^2+PC^2 = GA^2+GB^2+GC^2 + 3PG^2` for any point `P`.
- **Reflections in the plane (2D):** A reflection mapping `a -> b` has axis along `a + b` 
(angle bisector), with `|a|=|b|`. Normalize axis to unit `u=(u_1,u_2)`. Reflection 
matrix:
  `R = \begin{pmatrix} u_1^2-u_2^2 & 2u_1u_2 \\ 2u_1u_2 & u_2^2-u_1^2 \end{pmatrix}`.
  Or reflect `v` via `2(v*u)u - v`.
  Example: reflection `(5,0)->(4,3)`; axis `proportional-to(9,3)proportional-to(3,1)`, `u=\frac{1}{\sqrt{10}}(3,1)`; 
reflecting `(-2,3)` gives `\begin{pmatrix} 1/5 \\ -18/5 \end{pmatrix}`.

Always end with the final answer inside `\boxed{}`, containing only the exact answer.
\end{Verbatim}

\subsection{Qwen v2 SkillOpt skill, MATH-500 (condition C, full text)}

SHA-256:
\hashvalue{33225df46b7d6fec627d019480cab1a402d757e45fa398149d8c2aa9bdad68d3}.
\begin{Verbatim}
# Skill

Solve the math problem. Give the final answer inside \boxed{}.

## Response Guidelines
- Always produce output. Begin reasoning immediately and keep it concise; do not stall.
- Bound your work: aim for a short, direct derivation rather than exhaustive 
exploration. Avoid extremely long or repetitive chains that risk timing out.
- ALWAYS end with the final answer inside \boxed{}. Never leave the response empty.
- Match the answer's format to what is asked (e.g., a fraction like 
\boxed{\frac{3}{56}}, an exact expression like \boxed{3R^2}, or a dollar amount).

- If the problem asks for an angle 'in degrees', include the degree symbol in the boxed 
answer (e.g., \boxed{90^\circ}, \boxed{145^\circ}), not just the bare number. Likewise 
attach any requested units. When the answer is a degree measure, prefer matching the 
gold format with ^\circ.

## Approach
1. Translate the problem into explicit mathematical relationships (equations, 
inequalities, constraints); define variables/substitutions early to simplify before 
manipulating.
2. Work in small, clearly labeled steps (distribute, combine like terms, isolate, 
substitute), showing each intermediate result.
3. Apply relevant identities, formulas, and shortcuts instead of brute force (e.g., 
binomial symmetry, complementary counting, Cauchy-Schwarz, angle-sum/polygon-angle 
formulas); reduce known relations (like x^2 = x+1) to express higher terms.
4. Use all given constraints (domain bounds, convexity, sign) to select the single 
valid answer among candidates. For maximum/minimum problems, after finding the bound, 
verify it is attainable by exhibiting values that achieve it.
5. Verify the result by substituting back or a quick sanity check before finalizing.
6. Report the answer in exact, fully simplified form (reduced fractions like 
\frac{a}{b}, exact values keeping symbols like \pi, ordered pairs) inside \boxed{}, 
matching the requested format.

## Efficient Techniques
- Polynomial interpolation / values at extra points: use finite differences or the fact 
that a degree-n polynomial has constant n-th differences; construct an auxiliary 
polynomial that vanishes at the given nodes to avoid heavy computation.
- Triangle geometry (centroid, orthocenter, circumradius): use standard 
vector/coordinate identities such as sum of squared distances from a point P: \sum PA^2 
= \sum GA^2 + 3\,PG^2, and known relations OH, OG, a^2+b^2+c^2 = ... to express results 
in a,b,c,R quickly.
- Percentage/word problems: translate directly into one equation and solve; e.g., '60% 
off' means sale price = 40% of original.

- Integer identification / floor & bounding problems: when an exact closed form is 
messy, pin the integer answer by trapping the target between consecutive integers with 
simple rational estimates or a double inequality (e.g., show 3 < r^2+r+1 < 4, or use 0 
< angle < 180 to fix a unique n). For 'which is largest/smallest' comparisons, just 
compute the deciding quantity (like d/t) for each candidate and compare directly.
- Roots of unity / cyclic powers: if z^n = 1 (z!=1), then 1+z+z^2+...+z^{n-1}=0 and 1/z^k = 
z^{n-k}. Rewrite reciprocal or negative powers as positive powers to collapse sums 
quickly.
- 'Smallest value' word problems: convert to a congruence (e.g., total equivalent-to 0 mod k for 
equal redistribution), combine with any inequality bound (e.g., total > 200), then take 
the least integer satisfying both. Reduce coefficients mod k early (7 equivalent-to -1 mod 8).
- Grid/collinearity counting: count total sets as C(N,k), then enumerate collinear 
configurations by line type (rows, columns, diagonals) and divide.

- Sector/annulus area problems: the area of a sector of angle \theta (degrees) in a 
circle of radius r is \frac{\theta}{360}\pi r^2; an annular sector is the difference of 
two such sectors. Set up one equation relating the shaded fraction to \theta and solve 
directly instead of exhaustive case analysis.
\end{Verbatim}

\subsection{Qwen v1 sensitivity evidence}

The v1 held-out files, analysis outputs, original pre-recovery snapshot, and
recovery log remain in the evidence archive. The pre-recovery MATH file has
SHA-256 (concatenate the two groups):
\texttt{81e759425c7266373bf1bbfb72024ebb}\allowbreak
\texttt{de64340a359fe549e2728c7ea35cb062}.
It records 945 unique rows before exactly 55 circuit-breaker-skipped manual
tasks were evaluated after endpoint restoration. The final v1 file preserves
the other rows and reports the remaining failures. These records document the
deviation and are not pooled with v2. Table~\ref{tab:qwen-v1-v2} reports the
complete-pipeline sensitivity results and whether each optimized artifact was
reselected identically.

\section{Reproduction}
\label{app:repro}

{\sloppy Software versions used throughout: Python 3.10.0; \texttt{gepa} 0.1.4
(official package); \texttt{skillopt} 0.2.0 (official package, installed
from a main-branch snapshot of the GitHub repository downloaded in August
2026; the snapshot carries no Git metadata, but its exact retained source tree
is identified below by a deterministic manifest hash); DwarfStar
(ds4, built from \url{https://github.com/antirez/ds4}) serving the
\texttt{ds4f-q2} checkpoint, pinned by SHA-256 in \S\ref{sec:hardware}.
The Qwen replication uses the same GEPA and SkillOpt versions, Qwen3.8-27B
MLX 8-bit served by LM Studio 0.4.14+4, and the frozen protocol at
\texttt{QWEN\_EXPERIMENT\_PROTOCOL.md}. Exact dataset, prompt, model, run,
and artifact hashes are retained under
\texttt{experiments/qwen38\_v2/manifests/} and each benchmark's
\texttt{manifests/} directory; corresponding v1 records are retained for the
sensitivity analysis.\par}

{\sloppy The exact retained SkillOpt 0.2.0 source snapshot contains 536 files
and is identified by the deterministic manifest
\texttt{manifests/skillopt\_vendored.jsonl}, SHA-256
\hashvalue{dd60eb6b42eec44fd64f839d57a995ee15cb576f658d406d34ffd82b4a0ece5d}.
The snapshot lacks Git metadata, and no upstream commit or source-archive hash
is recoverable. The Qwen checkpoint is likewise identified by per-file hashes;
the upstream revision and MLX conversion revision were not recorded.\par}

The framework is driven by one YAML config per experiment. The full pipeline
for each benchmark is three commands. These commands document the exact
historical invocation; the \texttt{gepa-skillopt-bench} implementation itself
is not included in the public evidence archive.

{\small
\begin{verbatim}
# HotpotQA, minimal-seed study
gepa-skillopt-bench --config config/benchmark_hotpotqa_minimal.yaml \
    --target deepseek_v4_flash_local optimize
gepa-skillopt-bench --config config/benchmark_hotpotqa_minimal.yaml \
    --target deepseek_v4_flash_local evaluate --split test
gepa-skillopt-bench --config config/benchmark_hotpotqa_minimal.yaml report

# MATH-500, minimal-seed study
gepa-skillopt-bench --config config/benchmark_math500_minimal.yaml \
    --target deepseek_v4_flash_local optimize
gepa-skillopt-bench --config config/benchmark_math500_minimal.yaml \
    --target deepseek_v4_flash_local evaluate --split test
gepa-skillopt-bench --config config/benchmark_math500_minimal.yaml report
\end{verbatim}
}

The Qwen substitution uses the two configurations
\begin{quote}\small
\path{config/benchmark_hotpotqa_qwen38_27b_v2.yaml}\\
\path{config/benchmark_math500_qwen38_27b_v2.yaml}.
\end{quote}
Its frozen execution order,
preflight requirements, freeze commands, and held-out evaluation contract are
specified in \path{QWEN_EXPERIMENT_PROTOCOL.md}. Because the reported run is
complete and frozen, these paths document reproduction rather than authorize
resuming or retuning it.

Datasets are fetched and split by \texttt{scripts/fetch\_benchmarks.py}
(fixed seed 42; sequential splits). Key config excerpt (MATH-500):

{\small
\begin{verbatim}
optimization:
  gepa:
    engine: official
    max_metric_calls: 400
    reflection_minibatch_size: 5
    reflection: {model: anthropic/claude-opus-4.8, temperature: 1.0}
  skillopt:
    engine: official
    epochs: 2
    batch_size: 15
    textual_learning_rate: 4
    selection_set_size: 50
    gate_metric: hard
    reflection: {model: anthropic/claude-opus-4.8, temperature: 0.2}
targets:
  - id: deepseek_v4_flash_local
    adapter: openai_compatible
    model: deepseek-v4-flash
    temperature: 0.0
    max_tokens: 4096
    timeout_seconds: 480
\end{verbatim}
}

{\raggedright Every table and figure is generated by
versioned analysis scripts from retained per-task outputs, aggregate
statistics, optimizer telemetry, and artifact manifests. The public archive
contains their resulting figure and table inputs plus a transformation ledger,
but excludes benchmark implementation code. The Qwen v1 recovery deviation,
original hash, and no-retry rule are retained with the sensitivity evidence;
the primary v2 run required no recovery.\par}

\typeout{get arXiv to do 4 passes: Label(s) may have changed. Rerun}
\end{document}